%% file: main.tex
\pdfoutput=1

\documentclass[11pt]{article}

\usepackage[final]{acl}

\usepackage{times}
\usepackage{latexsym}

\usepackage[T1]{fontenc}

\usepackage[utf8]{inputenc}

\usepackage{microtype}

\usepackage{inconsolata}

\usepackage{graphicx}

\input{math_commands}

\usepackage{url}
\usepackage{graphicx}
\usepackage{algorithm}
\usepackage{algpseudocode}
\usepackage{amsmath}
\usepackage{subcaption}
\usepackage{mathrsfs}
\usepackage{booktabs} 
\usepackage{todonotes}
\usepackage{tikz}
\usepackage{xcolor}
\usepackage{array}
\usepackage{colortbl} 

\title{Quantifying Compositionality of Classic and State-of-the-Art Embeddings}

\author{
  \textbf{Zhijin Guo}\textsuperscript{1,2}\thanks{\ \ Joint first author.} \quad
  \textbf{Chenhao Xue}\textsuperscript{1}\footnotemark[1] \quad
  \textbf{Zhaozhen Xu}\textsuperscript{2} \quad
  \textbf{Hongbo Bo}\textsuperscript{2} \quad
  \textbf{Yuxuan Ye}\textsuperscript{2} \\
  \textbf{Janet B. Pierrehumbert}\textsuperscript{1} \quad
  \textbf{Martha Lewis}\textsuperscript{3} \\
  \\
  \textsuperscript{1}University of Oxford \quad
  \textsuperscript{2}University of Bristol \quad
  \textsuperscript{3}University of Amsterdam \\
}

\begin{document}
\maketitle
\begin{abstract}

For language models to generalize correctly to novel expressions, it is critical that they exploit access compositional meanings when this is justified. Even if we don’t know what a “pelp” is, we can use our knowledge of numbers to understand that “ten pelps” makes more pelps than “two pelps”. Static word embeddings such as Word2vec made strong, indeed excessive, claims about compositionality. The SOTA generative, transformer models and graph models, however, go too far in the other direction by providing no real limits on shifts in meaning due to context. To quantify the additive compositionality, we formalize a two-step, generalized evaluation that (i) measures the linearity between known entity attributes and their embeddings via canonical correlation analysis, and (ii) evaluates additive generalization by reconstructing embeddings for unseen attribute combinations and checking reconstruction metrics such as L2 loss, cosine similarity, and retrieval accuracy. These metrics also capture failure cases where linear composition breaks down. Sentences, knowledge graphs, and word embeddings are evaluated and tracked the compositionality across all layers and training stages. Stronger compositional signals are observed in later training stages across data modalities, and in deeper layers of the transformer-based model before a decline at the top layer. Code is available at: \url{https://github.com/Zhijin-Guo1/quantifying-compositionality}.




\end{abstract}

\section{Introduction}\label{sec: introduction}

The representation of entities, concepts, and relations as vector embeddings is a foundational technique in machine learning \citep{mikolov2013efficient, mikolov2013distributed}. Despite their broad success, embeddings often lack interpretability. One approach to improving interpretability involves examining the \textit{compositionality} of embeddings: if embeddings can be understood as a function of known, interpretable representations, the information they encode becomes more accessible. 

Compositionality also enables models to understand and generate novel combinations of familiar components \citep{hofmann2025derivational}. For instance, in natural language, we can infer the meaning of a phrase like \textit{eco-friendly transport} even if we have never seen it before, simply by combining the meanings of \textit{eco-friendly} and \textit{transport}. In the context of embeddings, if a model can construct representations for unseen combinations by composing the embeddings of known parts, it suggests that the learned representation space is structured and generalizable.

Recent work has explored compositionality across word embeddings \citep{mikolov2013efficient}, sentence embeddings \citep{hewitt2019structural, xu2023on}, and graph embeddings \citep{bose2019compositional}, often highlighting examples like \textit{king} - \textit{man} + \textit{woman} $\approx$ \textit{queen} as evidence of linear semantic structure. However, these claims, particularly for static embeddings such as Word2Vec, have faced strong criticism. Notably, \citet{church2017word2vec} and others have argued that such findings are often cherry-picked and fail to generalize, thereby overstating the role of compositionality. In contrast, Transformer-based models such as BERT \citep{devlin2018bert}, GPT \citep{brown2020language}, and Llama \citep{touvron2023llama} offer more expressive and flexible architectures. These models typically rely on final-layer representations to encode sentence meaning, but often in ways that are less constrained by additive composition. In addition, many linguistic phenomena such as compound nouns (\textit{hot dog}), phrasal verbs (\textit{give up}), and sarcastic expressions (\textit{Great job, you broke it again}) cannot be accurately captured through additive mechanisms alone, as their meanings often depend on context rather than the meanings of individual components.

This contrast between early enthusiasm for additive structure and the complexity of meaning in modern models motivates a more rigorous investigation. It is essential to evaluate how much linear additive composition is actually retained across different architectures, modalities, and stages of training. Our work formalizes such an evaluation framework and applies it systematically to word, sentence, and graph embeddings. Specifically, we formalize a two-step, modality-agnostic evaluation: (i) measuring the linear alignment between known attributes and their embeddings using canonical correlation analysis (CCA), and (ii) assessing additive generalization by reconstructing embeddings for unseen attribute combinations and checking distance loss, cosine similarity, and retrieval accuracy.


Our study makes several novel contributions:
\begin{itemize}
\item Quantifying linearity and generalization: We formalize a unified evaluation pipeline by integrating canonical correlation analysis (CCA) with additive reconstruction to assess compositional generalization.

\item Evaluation generalized across modalities: Our evaluation is applied to Transformer-based sentence embeddings (e.g., BERT, GPT, Llama), static embeddings (Word2Vec), and knowledge graph embeddings used in recommender systems. We track compositional signals across layers and training stages.

\item Empirical insights: Embeddings show increasing additive compositionality during training, with improved generalization in Multi-BERT and a 1.5× rise in attribute–embedding correlation in graph models. Word embeddings align with semantic and syntactic structure and decompose into roots and suffixes.

\item Showing compositional limits: We quantify failure cases where additive methods fall short, revealing residuals that signal non-linear structure and unexplained semantics, guiding future research.
\end{itemize}

\section{Related Works}\label{sec:Related Works}
The classical principle that “the meaning of a whole is built from its parts’’ underlies linguistic theory and modern distributional semantics, motivating decades of work on whether and how vector embeddings encode such structure. Moving beyond intuition, \citet{elmoznino2024complexity} define \emph{representational compositionality} information-theoretically: a representation is compositional if it can be recoded as discrete symbol strings whose semantics is delivered by a \emph{low-complexity} mapping. In this paper we instantiate, and quantitatively test, the \textbf{linear additive} case of that criterion across words, sentences and knowledge-graph entities.

\noindent \textbf{When compositionality fails.} Not all language is built compositionally: idiomatic compounds (\textit{hot dog}), phrasal verbs (\textit{take off}), sentiment of expressions (\textit{yeah right}), and sarcasm invert or ignore the literal sum of their parts \citep{JackendoffRay2011CitP,Rodríguez-PuentePaula2012TDoN,dankers2023non}. Early computational work detects this breakdown by treating non-compositional expressions as a \textit{statistical outlier}: unusual association or semantic-asymmetry scores flagged verb–object or noun compounds that resisted substitution tests
\citep{tapanainen1998idiomatic,lin1999automatic}. A parallel strand relied on syntactic and selectional-preference cues \citep{cook2007pulling,mccarthy2007detecting}, while neural phrase-embedding models merge frequent idioms into single atomic tokens \citep{mikolov2013distributed}. Non-compositionality also manifests in vector spaces: analogy tests misrank many compounds that contradict the conceptual priors of compositionality \citep{church2017word2vec}, and work detects systematic non-compositionality in multiword expressions \citep{yazdani2015learning}. Even if compositionality exists, the mapping of conceptual components could be nonlinear, as quadratic or polynomial projections, complexity-theoretic definitions, or dynamic, role-dependent nonlinear combinations \citep{yazdani2015learning,elmoznino2024complexity,wang2024composition}. The set of conceptual components could also be incomplete or defined inaccurately \citep{dalpiaz2018pinpointing,wei2008ontology}. These realities frame our scope: we analyze to what extent the conceptual field where compositionality presents and, we only test its linear-additive form defined over a fixed set of conceptual components.


\noindent \textbf{Linear structure across modalities.} Early word-analogy work established that semantic relations align with linear directions in Skip-gram vectors \cite{mikolov2013distributed,levy2014neural,gittens2017skip}; similar structures appear in word \citep{mikolov2013efficient,mikolov2013linguistic,goldberg2014word2vec,seonwoo2019additive,fournier2020analogies}, sentence embeddings \citep{bowman2016generating, hewitt2019structural, li2020sentence}, and representations of LLMs, vision–language models, and other deep learning models \citep{meng2022locating, hernandezlinearity,trager2023linear,lepori2023break}. \citet{park2024linear} formalise this “linear representation hypothesis”, tying linear probes to steerable directions via a causal inner product.

\noindent \textbf{Measuring compositionality.} Empirical work measures compositional structure in three main ways. (i) \emph{Geometry-based measures:} vector arithmetic in Skip-gram embeddings \citep{mikolov2013efficient}, formal additive-offset analyses \citep{gittens2017skip,seonwoo2019additive},
linear decomposition for subject, verb and object in a sentence \citep{xu2023on}, graded TRE approximation scores \citep{andreas2019measuring}, and higher-order ICA correlations \citep{oyama2024understanding} all ask whether representations decompose as (near)linear sums of parts. (ii) \emph{Probing and perturbation:} simple classifiers or representational-similarity scores reveal which linguistic factors a layer encodes
\citep{hewitt2019structural,ettinger2016probing,tenney2019you,chrupala2019correlating,lepori-mccoy-2020-picking}. (iii) \emph{Behavioral generalization:} model outputs on nonce or zero-shot combinations expose the underlying mechanism, e.g.\ analogy vs.\ rules in derivational morphology \citep{hofmann2025derivational}. Our contribution is a unified and statistically robust two-step diagnostic consisting of linearity quantification and additive generalization. It applies across word, sentence, and knowledge graph embeddings, and supports layer- and time-resolved analysis of compositional structure.

\begin{figure*}[t]
  \centering
  \includegraphics[width=0.75\textwidth]{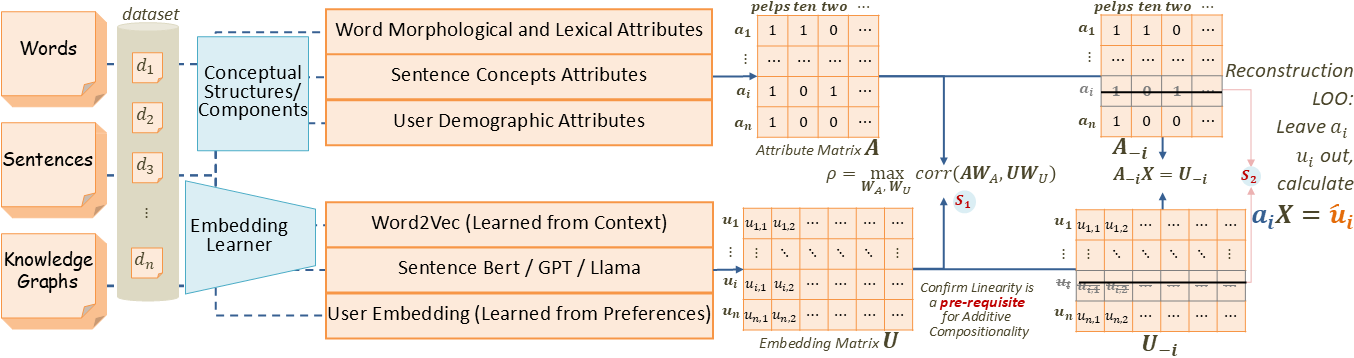}
  \caption{The pipeline of evaluating additive compositionality consists of quantifying linearity and generalizability.}
  \label{fig:tasks_workflow}
\end{figure*}

\section{Tasks and Formulations}

Previous work has explored methods to detect compositionality in embeddings by examining the relationships between entity attributes and their representations, including Linear or Syntactic Probing, Linear Representation Hypothesis, Correlation-Based Linearity Detection, Embedding of Labels and Additive Compositionality Detection~\citep{hewitt2019structural, thilaklidar, xu2023on, park2024linear, Guo2023}. Building on the geometric and behavioral strands surveyed in Section~\ref{sec:Related Works}, we formalize a two-step, modality-agnostic diagnostic to quantify compositional generalization. First assesses linearity between the attributes matrix and their embedding spaces. Second, tests whether those linear parts combine to predict embeddings for unseen attribute mixes. The pipeline is shown in Figure~\ref{fig:tasks_workflow}. For each entity, we consider two representations:


\begin{itemize}
    \item A \emph{binary attribute matrix} $\mathbf{A}$ (e.g., demographics or syntactic properties), where each row is an entity and each column represents an attribute. Entries $a_{ij} \in \{0, 1\}$ indicate the absence or presence of attribute $j$ in entity $i$.
    \item A \emph{continuous embedding matrix} $\mathbf{U}$ (e.g., user behaviour embeddings or word embeddings), where each row corresponds to an entity's embedding and each column represents a dimension of the embedding space.
\end{itemize}

\subsection{Assessing Linearity}
We employed the Canonical Correlation Analysis (CCA)~\citep{shawe2004kernel,xu2023on} to uncover latent correlations between between the binary attribute matrix $\mathbf{A}$ and the continuous embedding matrix $\mathbf{U}$.

Given the binary attribute matrix $\mathbf{A} \in \{0, 1\}^{q \times n}$ and the embedding matrix $\mathbf{U} \in \mathbb{R}^{q \times m}$, where $q$ is the number of entities, $n$ is the number of attributes, and $m$ is the dimensionality of the embeddings, the goal is to find transformation matrices $\mathbf{W}_A \in \mathbb{R}^{n \times k}$ and $\mathbf{W}_U \in \mathbb{R}^{m \times k}$ that maximize the correlation between the projected data:
$
\rho = \max_{\mathbf{W}_A,\, \mathbf{W}_U} \operatorname{corr}\left( \mathbf{A} \mathbf{W}_A,\, \mathbf{U} \mathbf{W}_U \right)
$.
Here, $\operatorname{corr}$ denotes the Pearson Correlation Coefficient (PCC) between the projected representations, and $k$ is the number of canonical components. The transformations $\mathbf{A} \mathbf{W}_A$ and $\mathbf{U} \mathbf{W}_U$ capture the most correlated aspects of the attributes and embeddings, respectively. We provide a detailed formal exposition of all settings necessary to define a CCA component, including the precise definition, the underlying vector space and mapping of semantic attributes, as well as illustrative examples. A complete derivation and examples are included in the Appendix \ref{sec: detail_cca}.

\subsection{Quantifying Additive Generalization}
An attribute embeddings matrix $\mathbf{X}$ quantifies how much each attribute influences the embeddings~\citep{kumar2019group, xu2023on}. It assumes that an entity's embedding can be approximated as a linear combination of attribute embeddings~\citep{seonwoo2019additive}. Given $\mathbf{A}$ and $\mathbf{U}$, the goal is to solve the linear system
$\mathbf{A} \mathbf{X} = \mathbf{U}$
where $\mathbf{X} \in \mathbb{R}^{n \times m}$ is the matrix of attribute embeddings to be learned.

We employed the Leave-One-Out (LOO) experiment to quantify the ability of inferring the embedding vector of an unseen entity's attributes. A LOO is performed for each entity $i$. We firstly exclude the $i$-th row from $\mathbf{A}$ and $\mathbf{U}$ to obtain $\mathbf{A}_{-i}$ and $\mathbf{U}_{-i}$, then solve $\mathbf{A}_{-i} \mathbf{X} = \mathbf{U}_{-i}$ using the pseudo-inverse to obtain $\mathbf{X}$. We then estimate the left-out embedding using $\hat{\mathbf{u}}_i = \mathbf{a}_i \mathbf{X}$, where $\mathbf{a}_i$ is the $i$-th row of $\mathbf{A}$, and finally compare $\hat{\mathbf{u}}_i$ with the actual embedding $\mathbf{u}_i$ using the following metrics: (more details are given in Appendix \ref{sec: loo})
\begin{enumerate}
   \item \emph{L2 Loss (Linear System Loss)}: $L_2 = \|\hat{\mathbf{u}}_i - \mathbf{u}_i\|^2$, measuring the reconstruction error.
   \item \emph{Embedding Prediction (Cosine Similarity)}: $\cos(\theta) = \dfrac{\hat{\mathbf{u}}_i \cdot \mathbf{u}_i}{\|\hat{\mathbf{u}}_i\| \, \|\mathbf{u}_i\|}$, assessing the alignment between the predicted and actual embeddings.
   \item \emph{Identity Prediction (Retrieval Accuracy)}: Determines if $\hat{\mathbf{u}}_i$ correctly identifies entity $i$ by checking if $\mathbf{u}_i$ is the nearest neighbor to $\hat{\mathbf{u}}_i$ among all embeddings.
\end{enumerate}

\paragraph{Hypothesis Testing}
To determine statistical significance, a non-parametric hypothesis test is performed by directly estimating the $p$-value through Monte Carlo sampling. First, the test statistic $T_{\text{real}}$ is computed for the real pairing: for CCA, this is the canonical correlation $\rho_{\text{real}}$; for additive compositionality, they are L2 loss, cosine similarity and retrieval accuracy based on the real pairing. Next, permuted pairings are generated by randomly shuffling the row order of $\mathbf{A}$ to disrupt the alignment, resulting in permuted datasets $\{\mathbf{A}^{(1)}, \mathbf{A}^{(2)}, \dots, \mathbf{A}^{(N)}\}$.

\section{Experiments}

\subsection{Dataset and Experiment Setup}
\noindent \textbf{Sentence} We evaluate the extent to which sentence embeddings from SBERT, GPT, and Llama can be additively decomposed into concepts expressed in the sentence, using a dataset consisting of sentences annotated with concepts, derived from the Schema-Guided Dialogue (SGD) dataset~\citep{rastogi2020towards}\footnote{https://github.com/google-research-datasets/dstc8-schema-guided-dialogue}. An example pair of sentence and concept labels is:
\textbf{Sentence:} Can you find me an \textcolor{teal}{Adventure} movie playing at \textcolor{blue}{AMC NewPark} in \textcolor{red}{Newark}?
\textbf{Concepts:} [\textit{\textcolor{red}{location}}, \textit{\textcolor{blue}{theater name}}, \textit{\textcolor{teal}{genre}}].

We select 2,458 sentences, each annotated with minimum 3 and maximum 4 concepts from a total set of 47 concepts. The mean number of concepts per sentence is 3.16. There are 90 unique combinations of concepts used. The sentences used comprise the test set of the Schema-Guided Dialogue (SGD) dataset. This split was chosen because it is annotated with denser labels, providing more comprehensive concept coverage.

We first look at the compositionality of sentence embeddings from the final layer of each model. We go on to look at how compositionality develops during the training stages of the MultiBERTs, and finally examine the compositionality of sentence embeddings through different layers of SBERT.

Embeddings are generated using pretrained sentence models, producing matrices $\mathbf{U} \in \mathbb{R}^{2,458 \times d}$, where $d$ is the dimensionality of the model. For SBERT\footnote{sentence-transformers/all-MiniLM-L6-v2} and Llama\footnote{meta-llama/Llama-2-7b-hf}, embeddings are generated by mean pooling over the token embeddings, excluding padding tokens. For GPT\footnote{text-embedding-3-small}, sentence embeddings are obtained from the OpenAI API, which provides precomputed embeddings representing the entire sentence. Each row of the matrix $\mathbf{U}$ consists of a sentence embedding for the corresponding sentence. We construct attribute matrices $\mathbf{A} \in \{0, 1\}^{2,458 \times 47}$ with each row being a binary vector indexing the relevant concepts for the corresponding sentence. In linearity quantification experiments, all sentences are used to compute correlations. In additive compositional generalization experiments, sentences are grouped by their concept combinations, and mean embeddings are computed for each group.

\noindent \textbf{Knowledge Graph} We use the MovieLens 1M dataset \citep{harper2015movielens} with 6040 users, 3900 movies, and 1 million ratings (1–5). Similar to sentence embedding, we pair each entity, i.e. user, with two descriptions: a binary vector representing demographic attributes (gender, age and occupation) and a user embedding learned from movie preferences (details in Appendix \ref{appendix:movielenstraining}).

For linearity quantification experiments, the continuous embedding matrix $\mathbf{U}_{corr} \in \mathbb{R}^{6040 \times 50}$ is populated with these embeddings. We also generate a binary attribute matrix $\mathbf{A}_{corr} \in \{0, 1\}^{6040 \times 9}$ containing one 9-dimensional binary attribute vector for each user, including attributes based on gender, age, and occupation. For additive compositional generalization experiments, we generate a binary attribute matrix $\mathbf{A}_{add} \in \{0, 1\}^{14 \times 9}$ containing one 9-dimensional binary attribute vector for each user including attributes based on gender and age (2 genders $\times$ 7 age groups). The continuous embedding matrix $\textbf{U}_{add} \in \mathbb{R}^{14 \times 50}$ is generated by taking the mean across embeddings associated with the same attribute combination.

\begin{table}[htbp]
    \centering
    \caption{Suffix presence (indicated by `1') for selected words from the MorphoLex dataset}
    \small
    \renewcommand{\arraystretch}{0.825}
    \begin{tabular}{p{1.6cm}rrrrrrr}
    \toprule
               Word &  er &  t &  y &  est &  ly & ness & less \\
    \midrule
      weightier &   1 &   1 &    1 &    0 &   0 &  0 &  0 \\
     weightiest &   0 &   1 &    1 &    1 &   0 &  0 &  0\\
       weightily &   0 &   1 &    1 &    0 &   1 &  0 &  0\\
        weightiness &   0 &   1 &    1 &    0 &   0 &  1 &  0\\
    weightlessly &   0 &   1 &    0 &    0 &   1 &  0 &  1\\
    \bottomrule
    \end{tabular}

    \label{tab:MorphLex}
\end{table}

\noindent \textbf{Word} We use WordNet embeddings as semantic representation, and MorphoLex for syntactic structures to examine semantic and syntactic information contained in word2vec embeddings. By comparing the word2vec embeddings against both WordNet and MorphoLex, we are able to disentangle the semantic and syntactic aspects of the word2vec representation. Training details for WordNet are provided in Appendix \ref{appendix: WordNet}. using vectors from GoogleNews-vectors-negative300, We build two continuous embedding matrices $\mathbf{U}_{w2v-wn}\in \mathbb{R}^{25,781\times 300}$ and $\mathbf{U}_{wn-w2v}\in \mathbb{R}^{25,781\times 20}$.

MorphoLex \citep{sanchez2018morpholex} provides a standardized morphological database derived from the English Lexicon Project, encompassing 68,624 words with nine variables for roots and affixes. An example extract from the dataset is given in Table \ref{tab:MorphLex}. In this paper, we focus specifically on words with one root and multiple suffixes. For linearity quantification experiments, 15,342 words and 81 suffixes are selected by intersecting the MorphoLex and word2vec (GoogleNews-vectors-negative300) vocabularies and filtering out suffixes occurring fewer than 10 times. We build a continuous embedding matrix $\mathbf{U}_{w2v-morpho}\in \mathbb{R}^{15,342\times 300}$ using vectors from GoogleNews-vectors-negative300 and a binary embedding matrix $\mathbf{X}_{morpho}\in \{0, 1\}^{15,342\times 81}$. 

To examine the additive compositionality of word2vec embeddings across multiple suffixes, we select 278 words as follows. We filter words that have exactly 3 suffixes, each of which occurs 10 times or more\footnote{These words are listed in Appendix \ref{appendix:278words}}. This results in 17 suffixes, but some words end up with fewer than 3 suffixes because certain suffixes occur less than 10 times and are filtered out. We build a continuous embedding matrix $\mathbf{U}_{w2v-add}\in \mathbb{R}^{278\times 300}$ using vectors from GoogleNews-vectors-negative300. We build a binary attribute matrix $\mathbf{A}_{morpho-add}\in \mathbb{R}^{278\times 45}$ indicating the presence or absence of each root word and morphemes.

\begin{figure}[t]
\centering
    \begin{subfigure}{0.475\columnwidth}
        \centering
        \includegraphics[width=\textwidth]{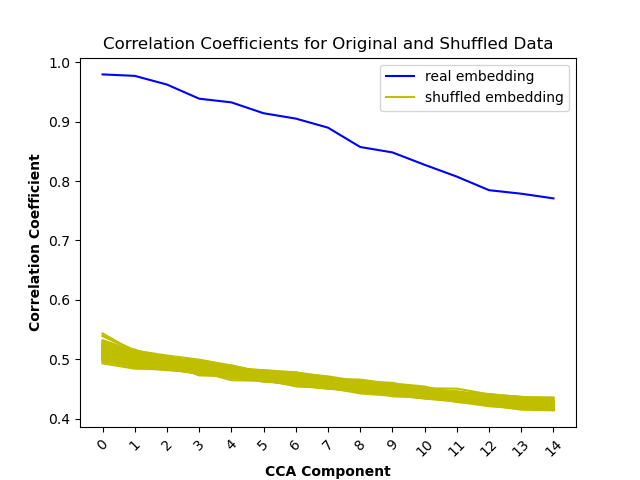}
        \caption{SBERT embeddings}
        \label{fig: permuted_sentence_CCA_bert}
    \end{subfigure}
    \hfill
    \begin{subfigure}{0.475\columnwidth}
        \centering
        \includegraphics[width=\textwidth]{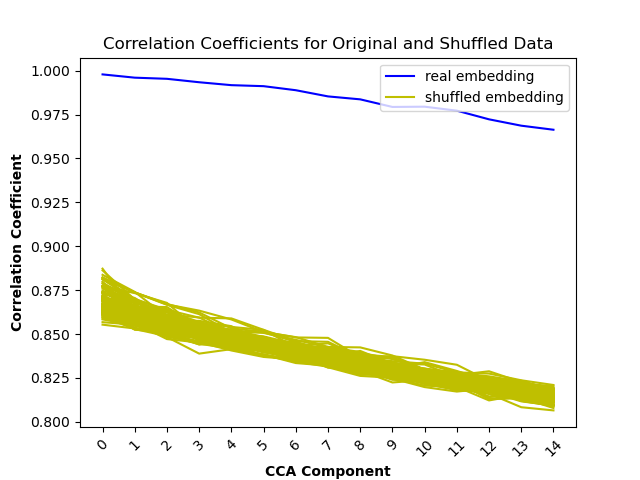}
        \caption{GPT Embeddings}
        \label{fig: permuted_sentence_CCA_gpt}
    \end{subfigure}
    \begin{subfigure}{0.475\columnwidth}
        \centering
        \includegraphics[width=\textwidth]{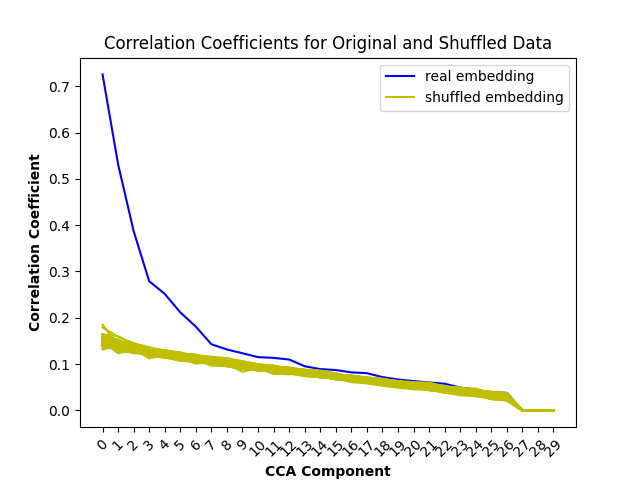}
        \caption{Embeddings by the additive scoring functions.}
        \label{fig: permuted_KG_CCA_TransE}
    \end{subfigure}
    \begin{subfigure}{0.475\columnwidth}
        \centering
        \includegraphics[width=0.925\textwidth]{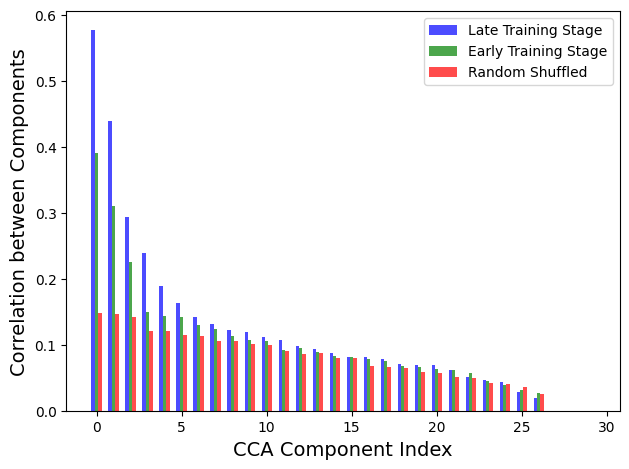}
        \caption{PCC over Training Stages of multiplicative model.}
        \label{fig: diff_training}
    \end{subfigure}
    \begin{subfigure}{0.475\columnwidth}
        \centering
        \includegraphics[width=\textwidth]{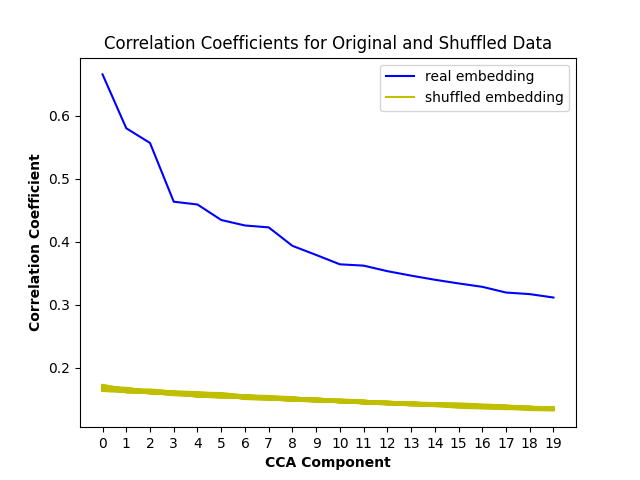}
        \caption{WordNet-word2vec}
        \label{fig: PCC_wordnet}
    \end{subfigure}
    \hfill
    \begin{subfigure}{0.475\columnwidth}
        \centering
        \includegraphics[width=\textwidth]{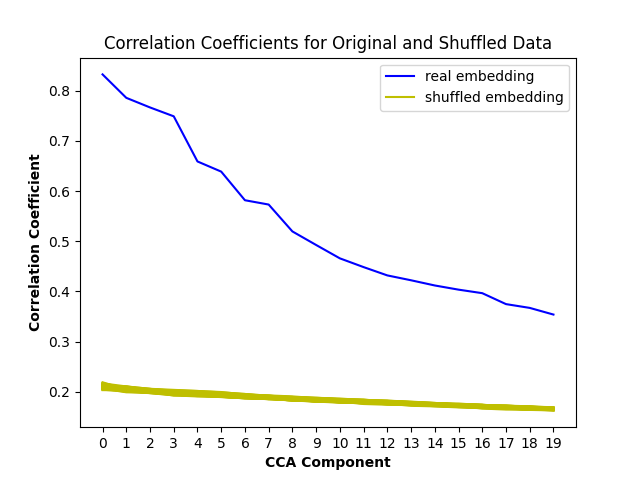}
        \caption{MorphoLex-word2vec}
        \label{fig: PCC_morphology}
    \end{subfigure}
    \caption{PCC (y-axis) between real (blue) and permuted (yellow) pairings across canonical components $k$ (x-axis). \textbf{Panels}: (a,b) sentence–concept (100 perms); (c) user–attribute (100 perms); (d) blue-late, green-early, red-permuted (e,f) 50 perms. PCC is computed between projected vectors $\mathbf{A}$ and $\mathbf{U}$; \textbf{real embedding always higher than the permuted baselines}.}
    \label{fig:permuted_sentence_CCA}
\end{figure}

\subsection{Quantifying Linearity}
\noindent \textbf{Sentence} For SBERT and GPT embeddings\footnote{Correlation for Llama embeddings is not assessed due to high dimensionality; see Limitations.}, we apply CCA to assess whether sentence embeddings correlate with their binary concept vectors. Figures \ref{fig: permuted_sentence_CCA_bert}, \ref{fig: permuted_sentence_CCA_gpt} show significant difference in correlation scores between real and permuted pairings (p-value $ < 0.01$) for both GPT and SBERT embeddings, indicating that sentence embeddings are correlated with their binary concept vectors.


\noindent \textbf{Knowledge Graph} Results of the correlation-based linearity analysis is reported in Figures \ref{fig: permuted_KG_CCA_TransE} for additive (TransE) scoring functions computed embeddings. As measured by CCA, the PCC for true user-attribute pairings substantially exceeds that for randomly permuted pairings, especially in the early canonical components (approximately three times). Moreover, we extract embeddings from both early (5 epochs) and late (300 epochs) stages of training to compare correlation-based linearity over time. The results show that the Pearson correlation coefficient (PCC) at the later stage is approximately 1.5 times higher than at the early stage.

\noindent \textbf{Word} As shown in Figures \ref{fig: PCC_wordnet} and \ref{fig: PCC_morphology}, the Pearson Correlation Coefficient (PCC) between word2vec, WordNet embeddings, and MorphoLex binary vectors substantially exceeds the randomized baseline, indicating that word2vec embeddings, trained on contextual co-occurrences, implicitly capture both semantic and morphological information. Results are reported in Figure \ref{fig: diff_training}. We see that as training progresses, the correlation coefficient increases, showing more demographic information becomes encoded into the embeddings. 

\begin{table}[htbp]
\centering
\caption{Generalizability metrics for SBERT, GPT, and Llama embeddings. Figures in \textbf{Real} columns are the mean across all leave-one out experiments. Figures in \textbf{Permuted} columns are the mean across 100 permutations of the sentence-concept pairs. All increases of Real over Permuted values are significant with $p$<0.01.}
\label{tab:compositionality_metrics}
\footnotesize
\begin{tabular}{p{1cm}|c ccc}
\hline
\textbf{Embed} & \textbf{Cosine} & \textbf{Cosine} & \textbf{Hits@5} & \textbf{Hits@5}  \\ 
              \textbf{-ding}     & \textbf{Real}  & \textbf{Permuted}       & \textbf{Real}    & \textbf{Permuted}      \\ \hline
\textbf{SBERT}      & 0.7761        & 0.4865            & 0.59            & 0.0405        \\            
\textbf{GPT}       & 0.7753        & 0.4941             & 0.57            & 0.0399         \\           
\textbf{Llama}     & 0.9355        & 0.8153             & 0.52            & 0.0468                  \\ \hline
\end{tabular}
\end{table}

\subsection{Quantifying Additive Generalization}
Once we confirm the conceptual components are linearly correlated with the embedding matrix, we can then quantify the additive generalizability.

\begin{figure*}[htbp]
    \centering
    \begin{subfigure}{0.275\textwidth}
        \centering
        \includegraphics[width=\textwidth]{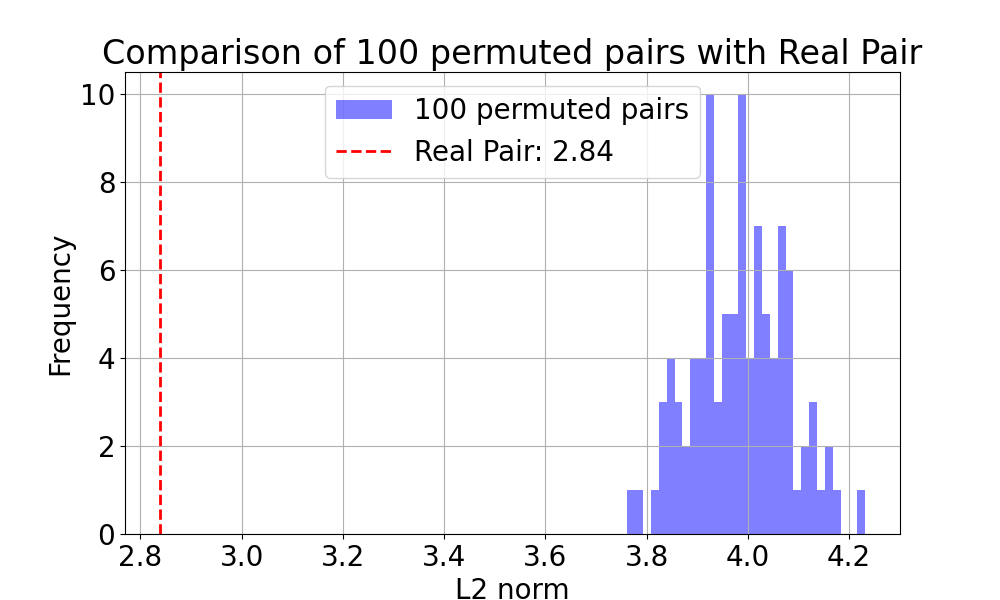}
        \centering
        \caption{L2 Loss {\scriptsize (SBERT)}}
        \label{fig: sentence_L2_bert}
    \end{subfigure}
    \begin{subfigure}{0.275\textwidth}
        \centering
        \includegraphics[width=\textwidth]{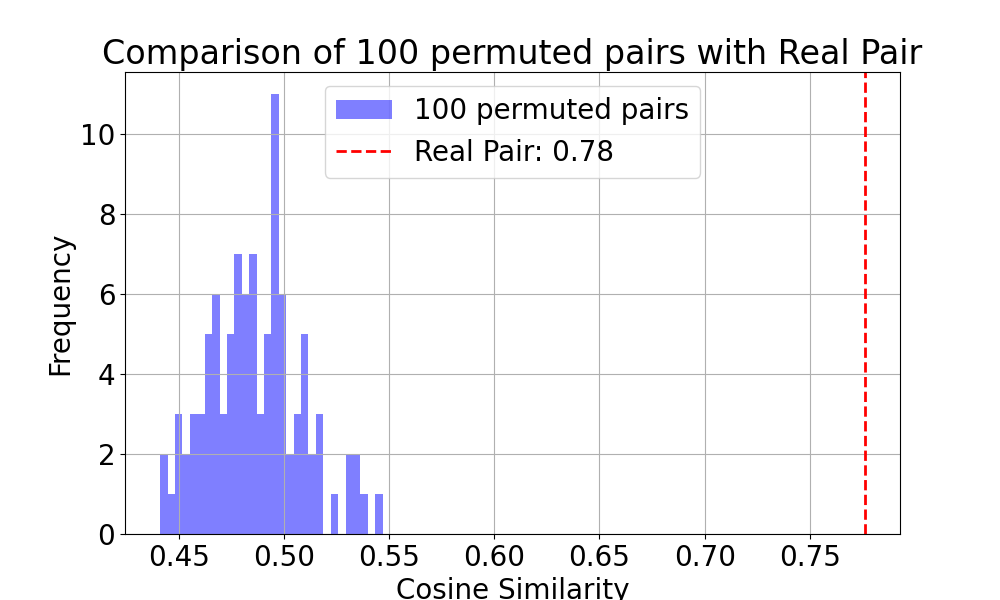}
        \caption{Cosine Similarity {\scriptsize (SBERT)}}
        \label{fig: sentence_cos_bert}
    \end{subfigure}
    \begin{subfigure}{0.275\textwidth}
        \centering
        \includegraphics[width=\textwidth]{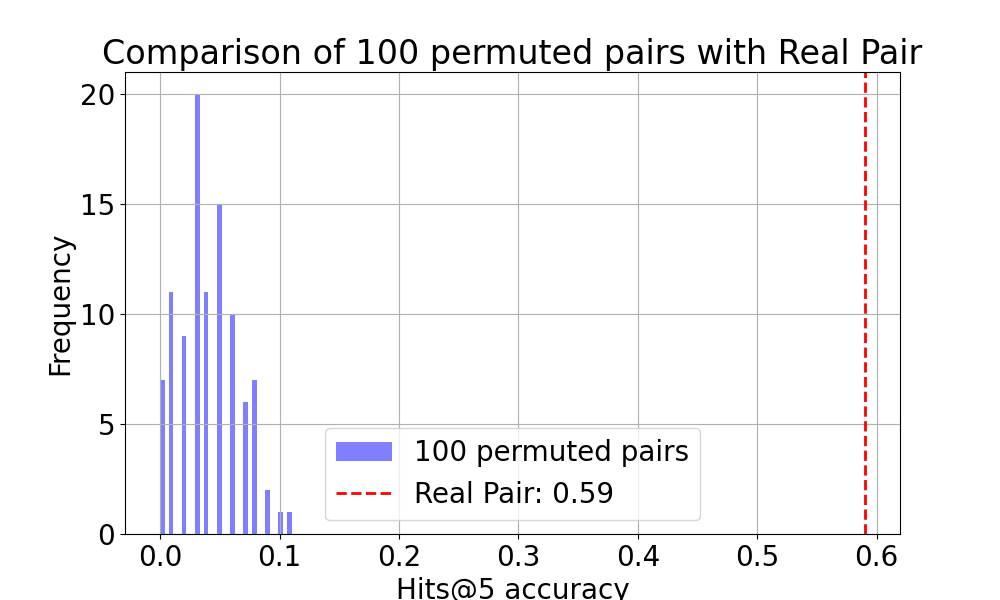}
        \caption{Retrieval Accuracy@5 {\scriptsize (SBERT)}}
        \label{fig: sentence_hits5_bert}
    \end{subfigure}
    \begin{subfigure}{0.275\textwidth}
        \centering
        \includegraphics[width=\textwidth]{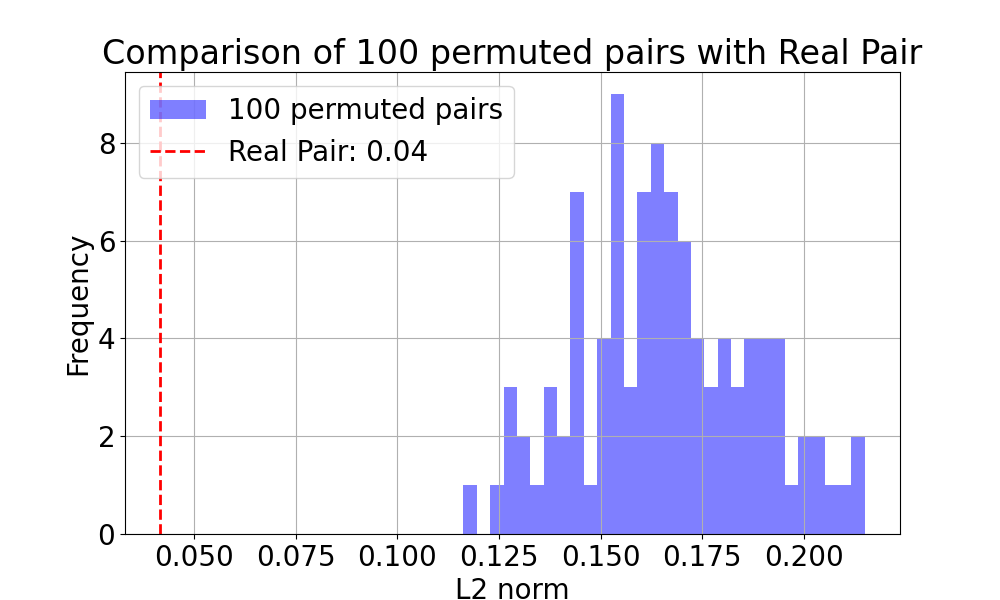}
        \centering
        \caption{L2 Loss {\scriptsize (KGE)}}
        \label{fig: KG_L2_TransE}
    \end{subfigure}
    \begin{subfigure}{0.275\textwidth}
        \centering
        \includegraphics[width=\textwidth]{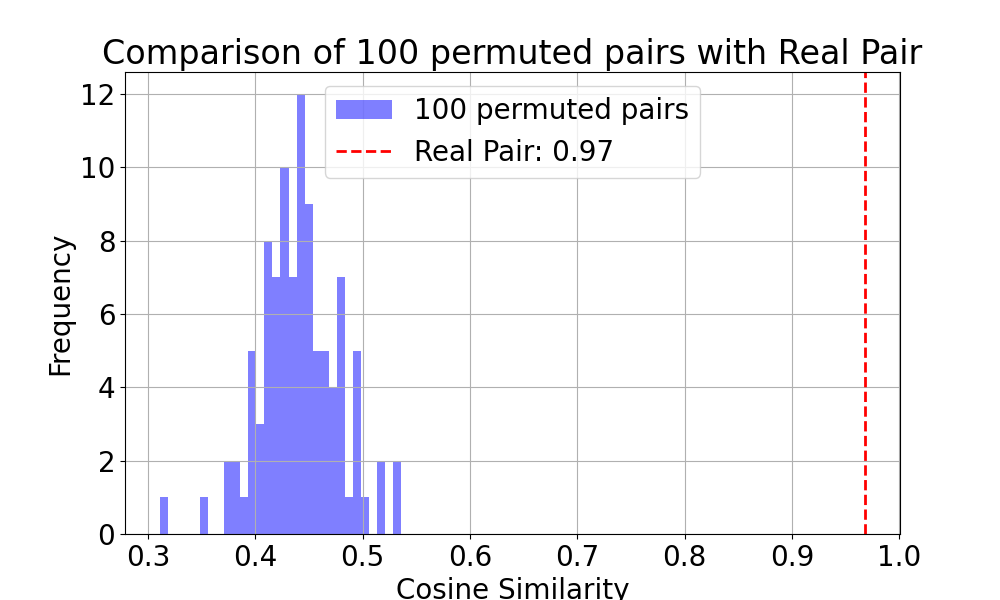}
        \caption{Cosine Similarity {\scriptsize (KGE)}}
        \label{fig: KG_cos_TransE}
    \end{subfigure}
    \begin{subfigure}{0.275\textwidth}
        \centering
        \includegraphics[width=\textwidth]{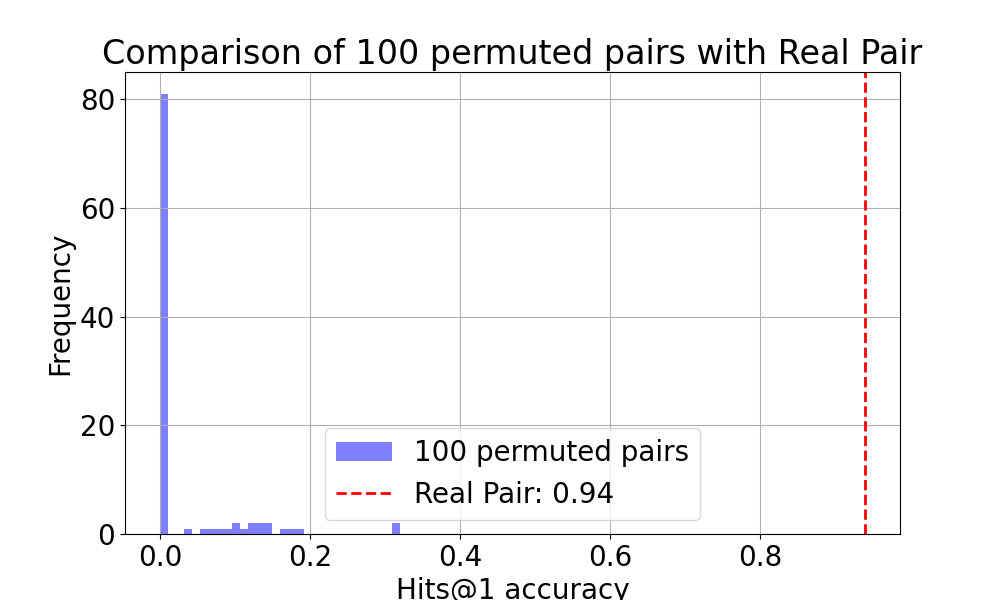}
        \caption{Retrieval Accuracy@1 {\scriptsize (KGE)}}
        \label{fig: KG_hits1_TransE}
    \end{subfigure}
    \begin{subfigure}{0.275\textwidth}
        \centering
        \includegraphics[width=\textwidth]{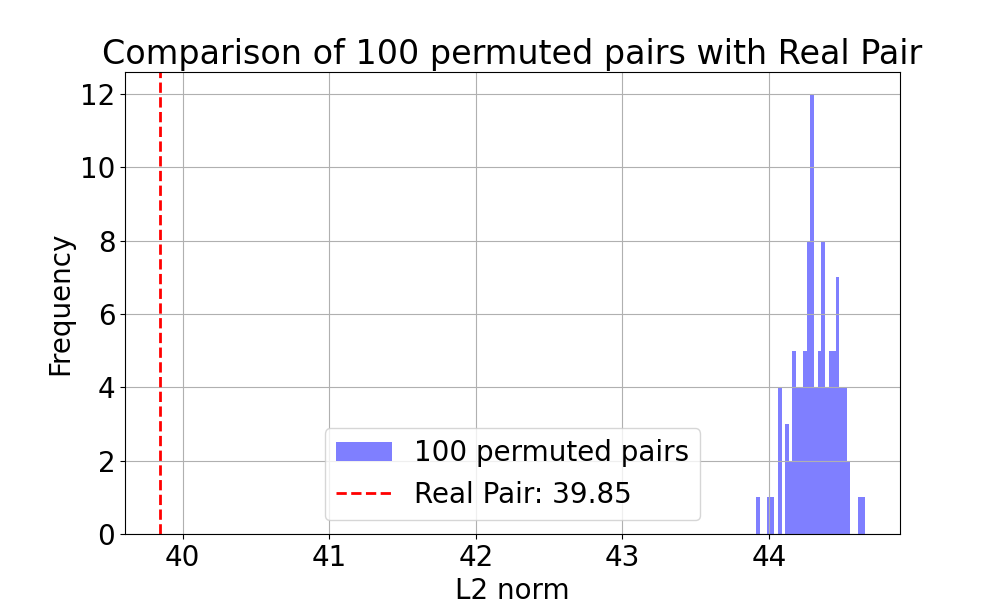}
        \centering
        \caption{L2 Loss {\scriptsize (word2vec)}}
        \label{fig: word2vec_L2}
    \end{subfigure}
    \begin{subfigure}{0.275\textwidth}
        \centering
        \includegraphics[width=\textwidth]{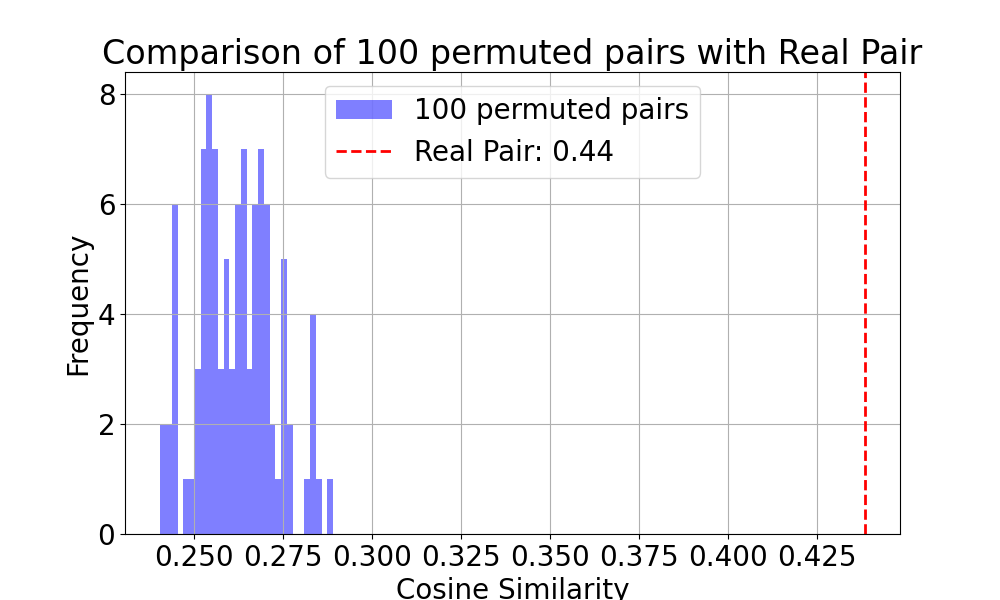}
        \caption{Cosine Similarity {\scriptsize (word2vec)}}
        \label{fig: word2vec_cos}
    \end{subfigure}
    \begin{subfigure}{0.275\textwidth}
        \centering
        \includegraphics[width=\textwidth]{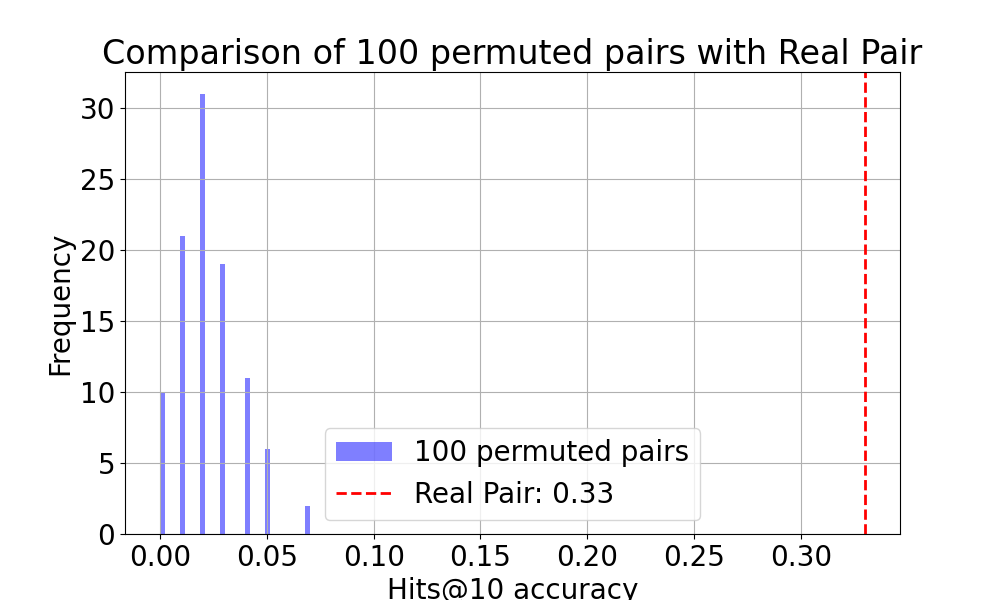}
        \caption{Retrieval Accuracy@10 {\tiny (word2vec)}}
        \label{fig: word2vec_hits10}
    \end{subfigure}
    \caption{\small Test statistics (y-axis) for embedding decompositions. Dashed line\,=\,mean scores of the predicted $\hat{\mathbf{u}}$ learned from real embedding pairs; bars\,=\,scores from 100 random permutations. Rows: SBERT, Knowledge Graph Embeddings, and word2vec. \textbf{Real pairings achieve higher cosine similarity and retrieval accuracy with lower L2 loss than permuted baselines}.}
    \label{fig:sentence_loo_results}
\end{figure*}

\noindent \textbf{Sentence} Results are reported in Table \ref{tab:compositionality_metrics}. Cosine similarity and retrieval accuracy for the true pairings are all significantly and substantially higher than for the permuted pairings, suggesting the additive compositionality in sentence embeddings can be generalized to unseen combination of concepts. The behaviour of SBERT and GPT embeddings are very similar, whereas Llama embeddings have a less substantial increase over permuted pairings in embedding prediction and retrieval accuracy (Hits@5). The distribution of the metric values for permuted pairs vs.\ real pairs for SBERT embeddings are plotted in Figures \ref{fig: sentence_L2_bert}, \ref{fig: sentence_cos_bert}, \ref{fig: sentence_hits5_bert}. Results for GPT and Llama embedding are provided in Appendix \ref{sec:sentence_embs}, Figures \ref{fig:sentence_loo_results_gpt} and \ref{fig:sentence_loo_results_llama}.

\noindent \textbf{Knowledge Graph} Results are reported in Figures \ref{fig: KG_L2_TransE}, \ref{fig: KG_cos_TransE}, \ref{fig: KG_hits1_TransE}, showing L2 loss of 0.04, cosine similarity of 0.97, and Hits@1 of 0.94, each outperforming the random baseline with $p = 0.01$. These findings reject the null hypothesis, demonstrating that user embeddings generalize additive relationships to new users' gender and age attributes.


\noindent \textbf{Word} Results are presented in Figures \ref{fig: word2vec_L2}, \ref{fig: word2vec_cos}, \ref{fig: word2vec_hits10}. We see that word2vec embeddings can be decomposed into root and multiple suffixes fairly well, and be generalized to unseen root-suffix combinations. The linear system loss is 39.85, lower than the minimum loss of the random system (43.91). Cosine similarity is 0.44, greater than all instances of the random baseline, and retrieval accuracy @10 is greater than that of the random system. However, overall these values are low, showing that there is still a fair bit of information that is not being captured by this representation.


\begin{table}[htbp]
\centering
\caption{Additive compositionality metrics across training steps of BERT. Number of training steps is given by the suffix to the model, e.g. cls\_20k indicates the performance of the model after 20k training steps. Rel. Diff expresses this difference as a percentage of the permuted similarity, showing proportional improvement. Figures in Real columns are the mean across all leave-one out experiments. Figures in Permuted columns are the mean across 100 permutations of the sentence-concept pairs.}
\label{tab:bert_training_stages}
\small
\begin{tabular}{l|p{2.15em}p{2.15em}p{2.15em}p{2.1em}p{2.25em}}
\hline
\textbf{Model} & \textbf{Cosine Real} & \textbf{Cosine Permuted} & \textbf{Cosine Rel. Diff} & \textbf{Hits @5 Real} & \textbf{L2 Loss Real} \\ \hline
\textbf{cls\_0k}     & 0.9884 & 0.9882  & 0.02\% & 0.44 & 29.39 \\ 
\textbf{cls\_20k}    & 0.8787 & 0.7722  & 13.79\% & 0.55 & 65.46 \\ 
\textbf{cls\_40k}    & 0.8773 & 0.7724  & 13.59\% & 0.48 & 62.48 \\ 
\textbf{cls\_100k}   & 0.9201 & 0.8323  & 10.54\% & 0.55 & 47.16 \\ 
\textbf{cls\_1000k}  & 0.9545 & 0.9149  & 4.33\% & 0.44 & 21.16 \\ 
\textbf{cls\_2000k}  & 0.9538 & 0.9094  & 4.89\% & 0.48 & 21.07 \\ \hline
\end{tabular}
\end{table}

\subsubsection{Comparison across Training Stages}
\noindent \textbf{Sentence} We further compare additive compositional generalization across different training stages of the MultiBERTs \citep{sellam2022multiberts}. We again use the same binary attribute matrix $\textbf{A}$ at each training stage. We build continuous embedding matrices $\mathbf{U}_{steps}$ for 0, 20k, 40k, 100k, 1000k, and 2000k training steps. We use the [CLS] token to represent sentences. As in Table \ref{tab:bert_training_stages}, at 0k steps, there is no substantial differences between the values of the cosine similarity metrics for real and permuted despite high overall values (0.9884). This indicates that the embeddings lack differentiation and do not capture conceptual relationships. After 20k training steps, generalization metrics (Cosine Real, L2 loss) constantly improve, showing generalization of additive compositionality emerges from training.

\subsubsection{Comparison between Different Layers}
\noindent \textbf{Sentence} We assess the additive compositional generalization of sentence embeddings derived from each layer of SBERT. At each layer, the binary attribute matrix $\textbf{A}$ remains the same as in the final layer experiment. Separate continuous embedding matrices $\textbf{U}_i$ are built for each layer $i$. To generate sentence embeddings at each layer, we extract hidden states for all layers from the SBERT model and apply the same pooling method as used for the last layer (mean pooling over token embeddings). Each embedding is normalized to length 1.

The results, reported in Table \ref{tab:Compositionality across Layers}, show that compositionality generalization increases through the layers, peaking at layer 4 or 5. However, an abrupt drop in compositionality generalization, as measured by cosine similarity, is observed at the last layer. This is in line with the phenomenon that semantic information is better encoded at earlier layers in the model, see for example experiments in the original BERT paper \citep{devlin2018bert}. Even though the cosine metric is fairly high for the random baseline, the discrimination between sentence meanings is still high as can be seen by fact that the Hits@5 for real pairings is consistently substantially higher than that for permuted pairings.

\section{Discussion and Conclusion}

\paragraph{Linearity and Compositionality}
Our paper adopted a two-step approach to explore the inherent structure of embeddings. In the first step, quantifying linearity confirmed a robust linear correlation between embeddings and candidate semantic features. This step established the necessary foundation for the second step, quantifying generalizability, where we rigorously demonstrated that final embeddings can be reconstructed as an additive combination of concept embeddings.

\begin{table}[t]
\centering
\caption{Additive compositionality metrics across SBERT layers. Figures in \textbf{Real} columns are the mean across all leave-one out experiments. Figures in \textbf{Permuted} columns are the mean across 100 permutations of the sentence-concept pairs. All Real vs. Permuted increases are significant at $p$ < 0.01.}
\small
\renewcommand{\arraystretch}{0.92}
\begin{tabular}{l|p{3em} p{3em}p{3em}p{3em}p{3em}p{3em}}
\hline
\textbf{Layer} & \textbf{Cosine Real} & \textbf{Cosine Permuted} & \textbf{Hits@5 Real} & \textbf{L2 Loss Real} \\ \hline
0 & 0.8889 & 0.7808 & 0.59 & 2.23 \\ 
1 & 0.9366 & 0.8671 & 0.57 & 1.80 \\ 
2 & 0.9397 & 0.8576 & 0.61 & 1.76 \\ 
3 & 0.9403 & 0.8330 & 0.64 & 1.73 \\ 
4 & 0.9408 & 0.8298 & 0.66 & 1.72 \\ 
5 & 0.9409 & 0.8273 & 0.62 & 1.71 \\
6 & 0.7761 & 0.4865 & 0.59 & 2.84 \\ \hline
\end{tabular}
\label{tab:Compositionality across Layers}
\end{table}

\begin{figure}[t]
    \centering
    \includegraphics[width=0.75\columnwidth]{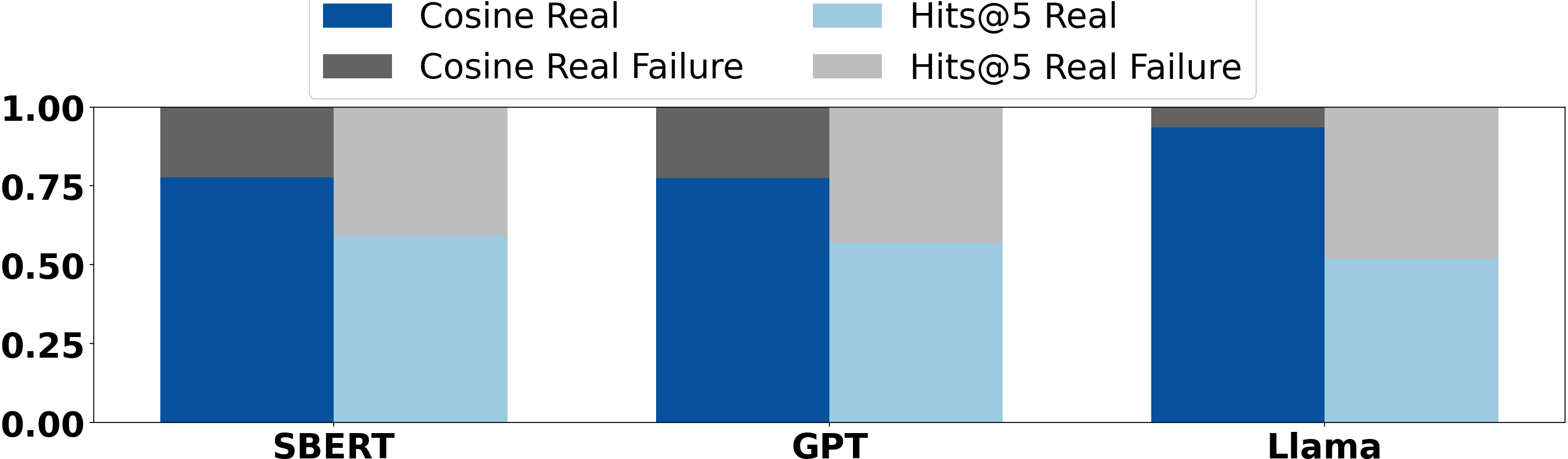}
     \caption{\small Comparison between SBERT, GPT, Llama Failures in Compositionality, failures are the cases that embeddings cannot be decomposed into components or vice versa. }
     \label{fig: Sentence_AddComp_single}
\end{figure}

Our findings address the broader question of whether relations between structured entities can be captured through simple vector operations. Evaluating a range of word, sentence, and graph embeddings showed a common thread of additive compositionality and therefore shows the generalizability. For sentence embeddings, conceptual components can often be combined additively, indicating that even large, non-linear models (e.g., GPT or Llama) retain interpretable vector structures. Word embeddings effectively encode both morphological and semantic relationships, as illustrated by transformations like \emph{weight} + \emph{y} + \emph{ly} = \emph{weightily}.

\paragraph{Quantifying Compositionality}The degree of compositionality increases across models regardless of training strategy. In MultiBERTs, compositionality steadily increase with training (see Cosine Real and L2 Norm in Table~\ref{tab:bert_training_stages}). In knowledge graph embeddings, the trends are similar. The high “Cosine Real score” observed with the initial, untrained embeddings (0k model) is noteworthy; however, it can be understood as a consequence of the model's initialization. This is because there is no significant statistical difference when comparing the “real pair” scores to those from “permuted pairs,” which serve as a baseline. On the other hand, in SBERT, compositionality increases in earlier layers but declines slightly in later layers, presumably because these upper layers specialize in task-specific representations \citep{devlin2018bert,tenney2019}. 

\paragraph{Understanding Compositional Failures}
Our framework not only detects successful instances of additive compositionality but also quantifies failure cases where reconstructions significantly diverge from the ground truth. These residuals highlight the limits of linear compositionality and are essential for identifying semantic phenomena, such as non-linear interactions or context-dependent meanings that cannot be captured by known attributes. For instance, we observed fluctuations in the Hits@5 score across training stages and transformer layers. One contributing factor may be inaccuracies in the conceptual representation of sentences (e.g., a theater name inadvertently revealing location information) an inherent challenge when mapping natural language to structured concepts.

We also compared multiple sentence embedding models, all of which exhibit some degree of compositionality failure when evaluated against an idealized “perfect compositionality” scenario. In such a case, the model would achieve perfect accuracy, and additive compositions would result in no residuals. These findings, presented in Figure~\ref{fig: Sentence_AddComp_single} ,underscore the limitations of current models and motivate future research into more expressive, non-linear approaches to compositional representation.

Overall, these results highlight that embeddings from diverse sources retain a surprising degree of additive compositional structure, offering valuable opportunities for improving interpretability in representation learning. At the same time, our framework quantifies instances where both static word embedding models and state-of-the-art transformer models fail to be decomposed in a linear way, consistently leaving residuals that point to unresolved semantic complexities, which shows an important direction for future research.
\paragraph{Conclusion}

We formalized a unified evaluation pipeline that aligns attribute matrices with embedding spaces via CCA and measures additive generalization through leave-one-out reconstruction. Applied to word, sentence, and knowledge graph embeddings, our experiments show that embeddings exhibit additive compositionality and can generalize to novel attribute combinations, though they also reveal consistent failure cases across models where linear composition breaks down. Our analysis quantifies such failure, underscoring the limits of linear composition. In transformers, the additive compositionality peaks in mid-to-late layers but declines in the final layer due to task specialization. We will release our code to support future work on more expressive, non-additive representations.

\section*{Limitations}

\begin{enumerate}
\item Conceptual Structure: Conceptual structures (such as concepts in sentence embeddings or user demographic information in knowledge graph embeddings) are rooted in human understanding. Some of these involve latent or poorly defined components that are difficult to model. As a result, failures in additive compositionality may stem from various factors, including inherent non-linearity, missing components, or incomplete representations.
\item Linearity assumption: As mentioned above, our method concentrates on detecting linearity-based compositionality, more methods are needed to handle non-linear cases.
\item The curse of dimensionality: Due to the large size of LLama embeddings, we did not compute correlation metrics. In future work, we plan to apply dimensionality reduction techniques, such as those based on the Johnson–Lindenstrauss lemma, to preserve statistical properties while improving computational efficiency and stability.
\item Dataset coverage: Experiments focus on English corpora and two KG benchmarks (FB15k-237, MovieLens-1M).  Results may not transfer to morphologically rich languages or heterogeneous graphs.
\end{enumerate}

\section*{Acknowledgments}
We would like to express our gratitude to Professor Nello Cristianini for his invaluable guidance and insightful discussions throughout the development of this work. His thoughtful feedback and generous time in repeatedly discussing the research ideas have greatly shaped and improved this study.

\bibliography{custom}

\appendix

\section{Details of Correlation-Based Compositionality Detection Method}
\label{sec: detail_cca}
Canonical Correlation Analysis (CCA) is used to measure the correlation information between two multivariate random variables \citep{shawe2004kernel}. Just like the univariate correlation coefficient, it is estimated on the basis of two aligned samples of observations.

A matrix of binary-valued attribute embeddings, denoted as $\mathbf{A}$, is essentially a matrix representation where each row corresponds to a specific attribute and each column corresponds to an individual data point (such as a word, image, or user). The entries of the matrix can take only two values, typically 0 or 1, signifying the absence or presence of a particular attribute. For example, in the context of textual data, an attribute might represent whether a word is a noun or not, and the matrix would be populated with 1s (presence) and 0s (absence) accordingly.

On the other hand, a matrix of user embeddings, denoted as $\mathbf{U}$, is a matrix where each row represents an individual user, and each column represents a certain feature or dimension of the embedding space. These embeddings are continuous-valued vectors that capture the movie preference of the users. The values in this matrix are not constrained to binary values and can span a continuous range.

These paired random variables are often different descriptions of the same object, for example genetic and clinical information about a set of patients \citep{seoane2014canonical}, French and English translations of the same document \citep{NIPS2002_d5e2fbef}, and even two images of the same object from different angles \citep{guo2019canonical}.

In the example of viewers and movies, we use this method to compare two descriptions of users. One matrix is based on demographic information, which are indicated by Boolean vectors. The other matrix is based on their behaviour, which is computed by their movie ratings only.

Assuming we have a vector for an individual user's attribute embedding, denoted as:
\[
\mathbf{a} = \left( a_1, a_2, \dots, a_n \right)^T
\]
and a corresponding individual user's computed embedding:
\[
\mathbf{u} = \left( u_1, u_2, \dots, u_m \right)^T
\]
we aim to explore the correlation between these two representations across multiple users. Given a set of \( q \) users, we define \( \mathbf{A} \) as a \( q \times n \) matrix where each row corresponds to the attribute embeddings for a specific user, and \( \mathbf{U} \) as a \( q \times m \) matrix where each row represents the computed embedding of the same user. Here, \( n \) is the number of attributes, and \( m \) is the dimensionality of the user embeddings.

Canonical Correlation Analysis (CCA) is then employed to find the projection matrices \( \mathbf{W}_A \in \mathbb{R}^{n \times k} \) and \( \mathbf{W}_U \in \mathbb{R}^{m \times k} \) that maximise the correlation between the transformed representations of \( \mathbf{A} \) and \( \mathbf{U} \). Each projection matrix contains \( k \)-projection vectors, where \( k \) is the number of canonical components that depend on the eigenvalues of the covariance matrix. For each \( k \)-th canonical component, the projections of the attribute and user embeddings are given by:
\[
\mathbf{A}_k = \mathbf{A} \mathbf{w}_{a_k} \quad \text{and} \quad \mathbf{U}_k = \mathbf{U} \mathbf{w}_{u_k}
\]
where \( \mathbf{w}_{a_k} \in \mathbb{R}^n \) and \( \mathbf{w}_{u_k} \in \mathbb{R}^m \) are the \( k \)-th projection vectors from the matrices \( \mathbf{W}_A \) and \( \mathbf{W}_U \), respectively. \( \mathbf{A}_k \in \mathbb{R}^q \) represents the projection of the original attribute matrix \( \mathbf{A} \) (size \( q \times n \)) onto the \( k \)-th canonical direction, using the projection vector \( \mathbf{w}_{a_k} \in \mathbb{R}^n \). It results in a vector of size \( q \times 1 \) that contains the transformed values for each user for the \( k \)-th component. Similarly, \( \mathbf{U}_k \in \mathbb{R}^q \) represents the projection of the original user embedding matrix \( \mathbf{U} \) (size \( q \times m \)) onto the \( k \)-th canonical direction, using the projection vector \( \mathbf{w}_{u_k} \in \mathbb{R}^m \). It also results in a vector of size \( q \times 1 \) that contains the transformed values for each user for the \( k \)-th component.

The goal of CCA is to maximise the Pearson Correlation Coefficient (PCC) between these transformed representations, i.e., between \( \mathbf{A}_k \) and \( \mathbf{U}_k \), for each \( k \)-th canonical component. The correlation for the \( k \)-th canonical component, denoted as \( \rho_k \), is given by the formula:

{\footnotesize
\begin{equation}
\rho_k = \frac{\sum_{i=1}^q \left( \left( \mathbf{A}_k \right)_i - \mu_{\mathbf{A}_k} \right) \left( \left( \mathbf{U}_k \right)_i - \mu_{\mathbf{U}_k} \right)}{\sqrt{\sum_{i=1}^q \left( \left( \mathbf{A}_k \right)_i - \mu_{\mathbf{A}_k} \right)^2 } \sqrt{\sum_{i=1}^q \left( \left( \mathbf{U}_k \right)_i - \mu_{\mathbf{U}_k} \right)^2 }}
\label{eq: PCC}
\end{equation}}

where \( \mu_{\mathbf{A}_k} \) and \( \mu_{\mathbf{U}_k} \) are the means of the transformed attribute and user embeddings, respectively, and \( q \) is the number of users.

In matrix form, we can express this objective as maximizing the correlation between the transformed matrices \( \mathbf{A} \mathbf{W}_A \) and \( \mathbf{U} \mathbf{W}_U \). This is formalized as:
\begin{equation}
\rho = \max_{\mathbf{W}_A, \mathbf{W}_U} \operatorname{corr} \left( \mathbf{A} \mathbf{W}_A, \mathbf{U} \mathbf{W}_U \right)
\end{equation}
where the correlation is maximised across the projection matrices \( \mathbf{W}_A \) and \( \mathbf{W}_U \), and the result is a set of \( K \)-canonical correlations \( \rho_1, \rho_2, \dots, \rho_k \) that describe the relationship between the attribute embeddings and the user embeddings for the entire dataset.

Thus, by computing the canonical correlations \( \rho_K \) for each component, we obtain insights into how well the attribute embeddings and computed user embeddings are aligned in terms of their underlying structure.

Consider the case where we have 4 users. The attribute matrix \( \mathbf{A} \) has dimensions \( 4 \times 4 \), representing 4 users and 4 attributes, and the user embedding matrix \( \mathbf{U} \) has dimensions \( 4 \times 3 \), representing 4 users and 3 embedding dimensions. Thus:
- \( \mathbf{A} \in \mathbb{R}^{4 \times 4} \): the attribute matrix where each row represents the 4-dimensional attribute vector for a user.
- \( \mathbf{U} \in \mathbb{R}^{4 \times 3} \): the user embedding matrix where each row represents the 3-dimensional embedding vector for a user.

Canonical Correlation Analysis (CCA) is employed to find the projection matrices \( \mathbf{W}_A \in \mathbb{R}^{4 \times k} \) and \( \mathbf{W}_U \in \mathbb{R}^{3 \times k} \) that maximize the correlation between the transformed representations of \( \mathbf{A} \) and \( \mathbf{U} \). Each projection matrix contains \( k \)-projection vectors, where \( k \) is the number of canonical components that depend on the eigenvalues of the covariance matrix. For each \( k \)-th canonical component, the projections of the attribute and user embeddings are given by:
\[
\mathbf{A}_k = \mathbf{A} \mathbf{w}_{a_k} \quad \text{and} \quad \mathbf{U}_k = \mathbf{U} \mathbf{w}_{u_k}
\]
where \( \mathbf{w}_{a_k} \in \mathbb{R}^4 \) and \( \mathbf{w}_{u_k} \in \mathbb{R}^3 \) are the \( k \)-th projection vectors from the matrices \( \mathbf{W}_A \) and \( \mathbf{W}_U \), respectively.
\section{Leave-one-out experiment}
\label{sec: loo}
\begin{enumerate}
    \item \textbf{Leave Out}: Exclude the $i$-th row from $\mathbf{A}$ and $\mathbf{U}$ to obtain $\mathbf{A}_{-i}$ and $\mathbf{U}_{-i}$.
    \item \textbf{Train}: Solve $\mathbf{A}_{-i} \mathbf{X} = \mathbf{U}_{-i}$ using the pseudo-inverse to obtain $\mathbf{X}$.
    \item \textbf{Predict}: Estimate the left-out embedding using $\hat{\mathbf{u}}_i = \mathbf{a}_i \mathbf{X}$, where $\mathbf{a}_i$ is the attribute vector of entity $i$.
    \item \textbf{Evaluate}: Compare $\hat{\mathbf{u}}_i$ with the actual embedding $\mathbf{u}_i$ using the following metrics:
       \begin{enumerate}
           \item \emph{L2 Loss}: $L_2 = \|\hat{\mathbf{u}}_i - \mathbf{u}_i\|^2$, measuring the reconstruction error.
           \item \emph{Embedding Prediction (Cosine Similarity)}: $\cos(\theta) = \dfrac{\hat{\mathbf{u}}_i \cdot \mathbf{u}_i}{\|\hat{\mathbf{u}}_i\| \, \|\mathbf{u}_i\|}$, assessing the alignment between the predicted and actual embeddings.
           \item \emph{Identity Prediction (Retrieval Accuracy)}: Determines if $\hat{\mathbf{u}}_i$ correctly identifies entity $i$ by checking if $\mathbf{u}_i$ is the nearest neighbor to $\hat{\mathbf{u}}_i$ among all embeddings.
       \end{enumerate}
\end{enumerate}

\section{Knowledge Graph Embedding}
\paragraph{Knowledge Graph Embedding}
A \textit{graph} $G = (V, E)$ consists of a set of vertices $V$ with edges $E$ between pairs of vertices. In a \textit{knowledge graph}, the vertices $V$ represent entities in the real world, and the edges $E$ encode that some relation holds between a pair of vertices. As a running example, we consider the case where the vertices $V$ are a set of viewers and films, and the edges $E$ encode the fact that a viewer has rated a film.

Knowledge Graphs represent information in terms of entities (or nodes) and the relationships (or edges) between them. The specific relation $ r $ that exists between two entities is depicted as a directed edge, and this connection is represented by a triple $(h,r,t)$. In this structure, we distinguish between the two nodes involved: the \emph{head}  ($ h $) and the \emph{tail}  ($ t $), represented by vectors $ \mathbf{h} $ and $ \mathbf{t} $ respectively. Such a triple is termed a \emph{fact}, denoted by $ f $:
\[
f = (h,r,t)
\]

\paragraph{Embedding Function.}
A knowledge graph $G = (V, E)$ can be embedded by assigning each node $v \in V$ a vector $\mathbf{x}_v \in \mathbb{R}^d$. 
This embedding function, $\Phi_{KG}: V \rightarrow \mathbb{R}^d$, maps nodes into a continuous space where their relationships are captured by a scoring or distance function. 
For instance, one could define a threshold $\theta$ such that an edge $(v_i, v_j) \in E$ exists if and only if $D(\mathbf{x}_{v_i}, \mathbf{x}_{v_j}) < \theta$.  
Conversely, given a set of embedded points, links between nodes can be recovered by applying a learned scoring function $S\bigl(\mathbf{x}_{v_i}, \mathbf{x}_{v_j}\bigr)$.

\subsection{Multiplicative Scoring}
\citet{nickel2011three} introduced a tensor-factorization approach for relational learning, treating each frontal slice of a three-dimensional tensor as a co-occurrence matrix for a given relation. 
In this model, a triple $(h, r, t)$ is scored using embeddings $\mathbf{h}, \mathbf{R}, \mathbf{t}$, where $\mathbf{h}, \mathbf{t} \in \mathbb{R}^d$ represent head and tail entities, and $\mathbf{R} \in \mathbb{R}^{d \times d}$ represents the relation:
\begin{equation}
\label{eq:multiplicative}
S(h,r,t) \;=\; \mathbf{h}^{T}\,\mathbf{R}\,\mathbf{t}.
\end{equation}
Various specializations exist, such as DistMult \citep{yang2014embedding}, which restricts $\mathbf{R}$ to a diagonal matrix (reducing overfitting), and ComplEx \citep{trouillon2016complex}, which employs complex-valued embeddings for asymmetric relations. 
In this work, we adopt DistMult due to its simplicity and scalability, particularly its suitability for large knowledge graphs.

\subsection{Additive Scoring}
TransE \citep{bordes2013translating} introduces a translation-based perspective, where each relation is a vector that shifts the embedding of the head entity to the tail entity. 
A triple $(h, r, t)$ is scored by:
\begin{equation}
\label{eq: transe}
S(h,r,t) \;=\; \|\mathbf{h} + \mathbf{r} \;-\; \mathbf{t}\|,
\quad \mathbf{h}, \mathbf{r}, \mathbf{t} \;\in\; \mathbb{R}^d.
\end{equation}
For example, \emph{King} + \emph{FemaleOf} $\approx$ \emph{Queen}. This translation idea captures relational semantics by minimizing the distance between $\mathbf{h} + \mathbf{r}$ and $\mathbf{t}$.


\paragraph{Rating Prediction}

In alignment with \citep{berg2017graph}, we establish a function $P$ that, given a triple of embeddings $(\mathbf{h},\mathbf{R},\mathbf{t})$, calculates the probability of the relation against all potential alternatives.
{\scriptsize
\begin{eqnarray}
\tiny
P\left(\mathbf{h},\mathbf{R},\mathbf{t}\right)=\text{SoftArgmax}(S(f)) =\frac{e^{S(f)}}{e^{S(f)}+\sum_{r' \neq r\in \mathscr{R}} e^{S(f')}}
\label{eqn: loss}
\end{eqnarray}}
In the above formula, $f =(h,r,t)$ denotes a true triple, and $f'=(h,r',t)$ denotes a corrupted triple, that is a randomly generated one, that we use as a proxy for a negative example (a pair of nodes that are not connected). 

Assigning numerical values to relations $r$, the predicted relation is then just the expected value $
\text{prediction} = \sum_{r \in \mathscr{R}} r P\left(\mathbf{h},\mathbf{R},\mathbf{t}\right)$
In our application of viewers and movies, the set of relations $\mathscr{R}$ could be the possible ratings that a user can give a movie. The predicted rating is then the expected value of the ratings, given the probability distribution produced by the scoring function. $S(f)$ refers to the scoring function in \citet{yang2014embedding}.

To learn a graph embedding, we follow the setting of \citet{bose2019compositional} as follows, 

\begin{equation}
        L  = - \sum_{f \in \mathscr{F}} \log \frac{e^{S(f)}}{e^{S(f)}+\sum_{f' \in \mathscr{F}'} e^{S(f')}}
     \label{eq:actual_loss}
\end{equation}
This loss function maximizes the probabilities of true triples $(f)$ and minimizes the probability of triples with corrupted triples: $(f')$.

\paragraph{Evaluation Metrics}
We use 4 metrics to evaluate our performance on the link prediction task. 
These are root mean square error (RMSE, $\sqrt{\frac{1}{n} \sum_{i=1}^{n}\left(\hat{y}_{i}-y_{i}\right)^{2}}$, where $\hat{y}_i$ is our predicted relation and $y_i$ is the true relation), Hits@K - the probability that our target value is in the top $K$ predictions, mean rank (MR) - the average ranking of each prediction, and mean reciprocal rank (MRR) to evaluate our performance on the link prediction task. These are standard metrics in the knowledge graph embedding community.

\section{Additive Compositionality by Model}
\label{sec:sentence_embs}

\begin{figure}[htbp]
    \centering
    \begin{subfigure}{0.3255\columnwidth}
        \centering
        \includegraphics[width=\textwidth]{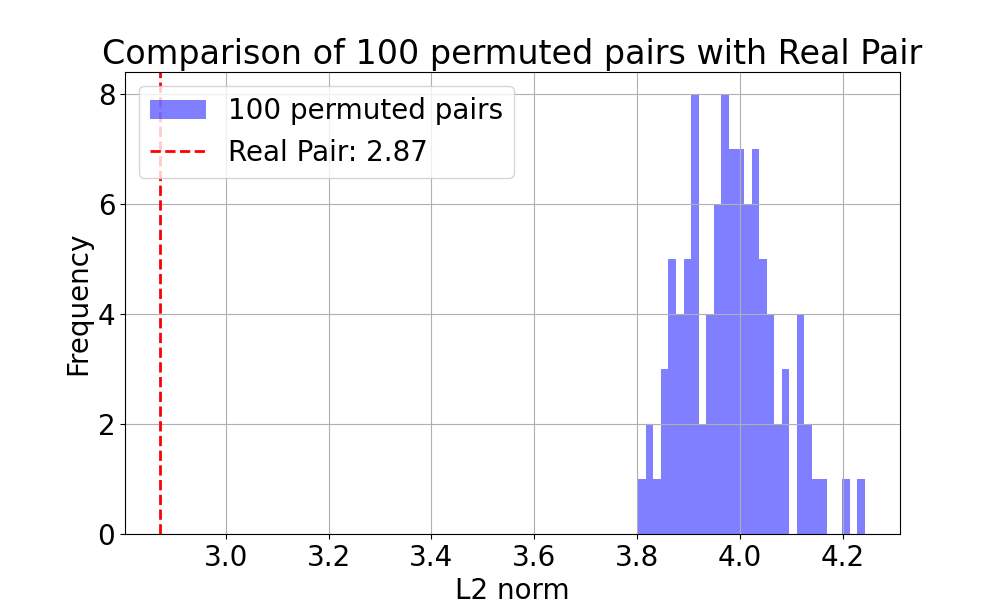}
        \centering
        \caption{Linear System Loss}
    \end{subfigure}
    \begin{subfigure}{0.3255\columnwidth}
        \centering
        \includegraphics[width=\textwidth]{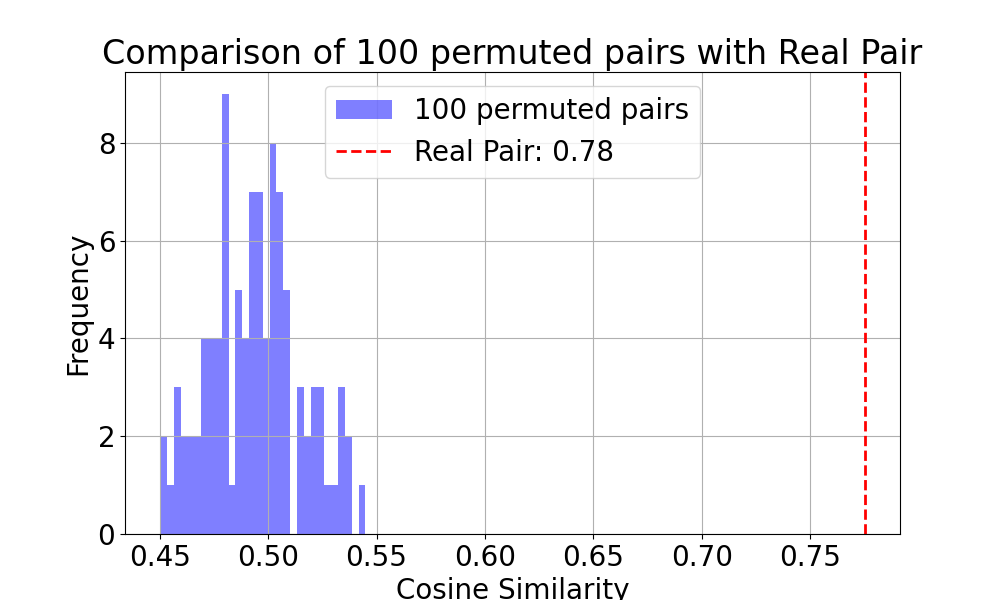}
        \caption{Cosine Similarity}
    \end{subfigure}
    \begin{subfigure}{0.3255\columnwidth}
        \centering
        \includegraphics[width=\textwidth]{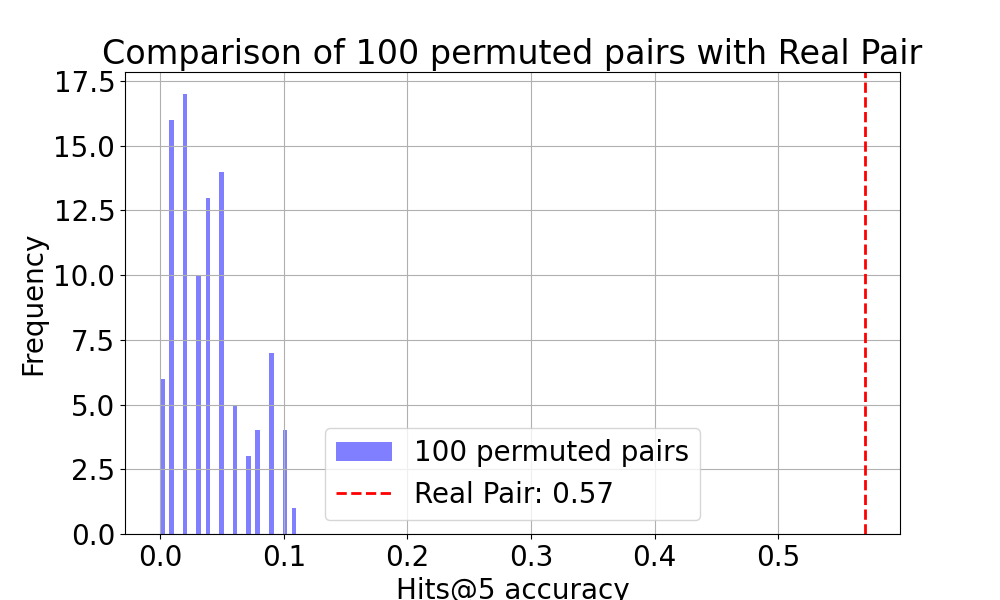}
        \caption{Retrieval Accuracy@5}
    \end{subfigure}
    \caption{Test statistics for GPT embedding decomposition. Dashed line is the average performance of $ \hat{\textbf{U}} $ learned from the user embedding. Bars are the distribution of the results from 100 random permutations.}
    \label{fig:sentence_loo_results_gpt}
\end{figure}

\begin{figure}[!h]
    \centering
    \begin{subfigure}{0.3255\columnwidth}
        \centering
        \includegraphics[width=\textwidth]{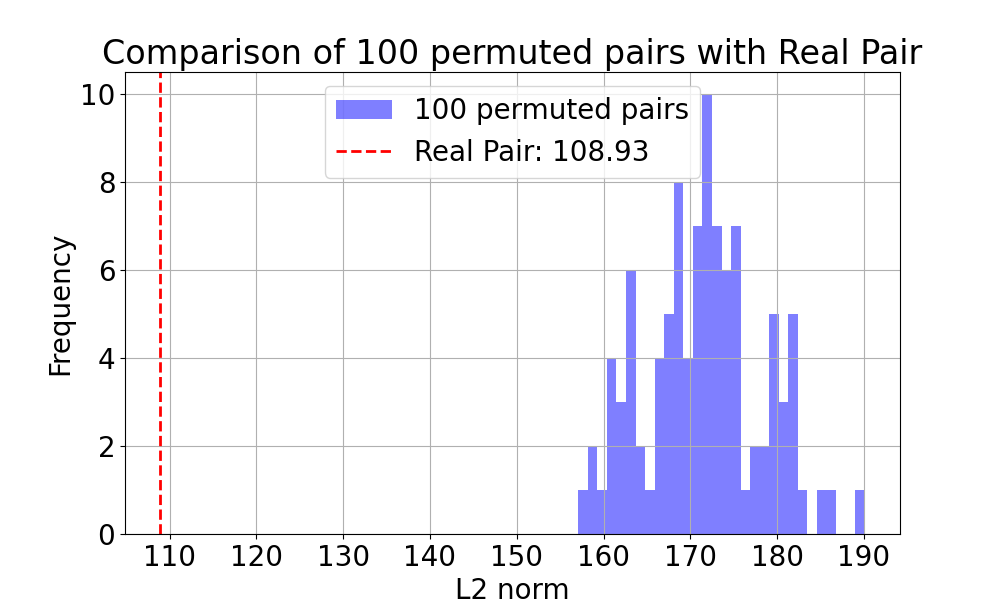}
        \centering
        \caption{Linear System Loss}
    \end{subfigure}
    \begin{subfigure}{0.3255\columnwidth}
        \centering
        \includegraphics[width=\textwidth]{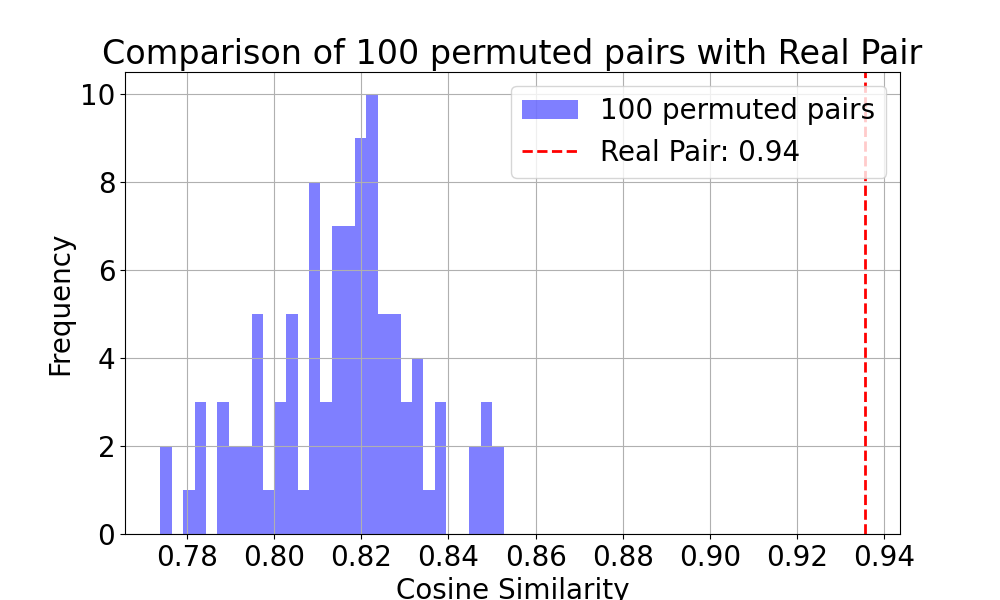}
        \caption{Cosine Similarity}
    \end{subfigure}
    \begin{subfigure}{0.3255\columnwidth}
        \centering
        \includegraphics[width=\textwidth]{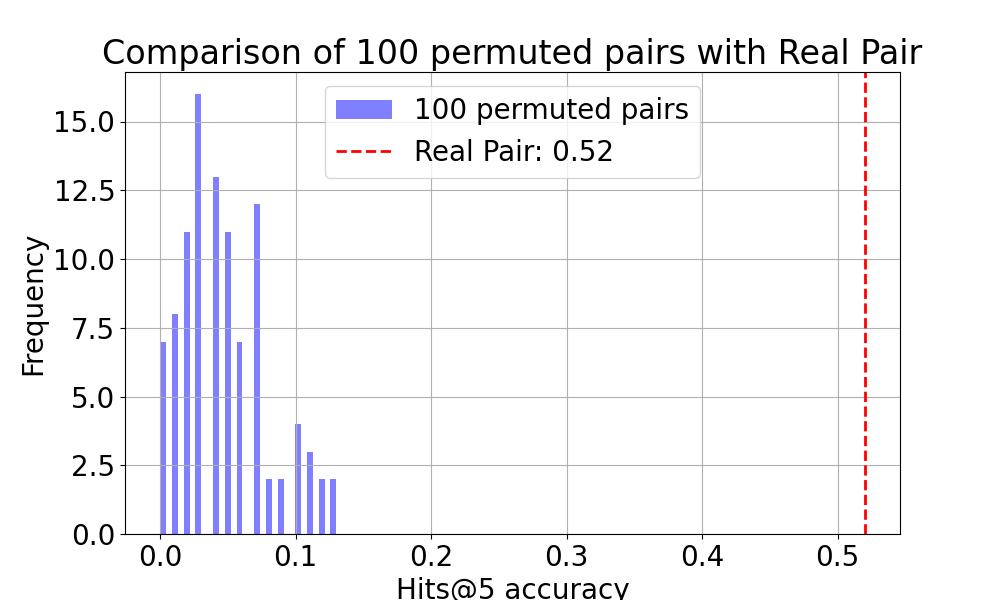}
        \caption{Retrieval Accuracy@5}
    \end{subfigure}
    \caption{Test statistics for Llama embedding decomposition. Dashed line is the average performance of $ \hat{\textbf{U}} $ learned from the user embedding. Bars are the distribution of the results from 100 random permutations.}
    \label{fig:sentence_loo_results_llama}
\end{figure}

\section{Compositionality across Layers and Training Stages}
\subsection{Comparison of Different Layers}
\paragraph{Comparison Metrics}
To fairly compare different layers, we cannot rely solely on raw cosine similarities or retrieval accuracies due to variations in scales and distributions. Instead, we use normalized metrics for comparability. The \textbf{Normalized Cosine Similarity} computes the difference between the mean real similarity and the mean permuted similarity, normalized by the maximum possible difference, $(1 - \text{Mean Permuted Similarity})$. The \textbf{Absolute Difference} is a simple measure of the difference between mean real and permuted similarities. Lastly, the \textbf{Relative Difference (Percentage Improvement)} expresses this difference as a percentage of the permuted similarity, indicating proportional improvement. These metrics enable robust and fair comparisons across models.
\begin{table*}[t]
\centering

\caption{Additive Compositionality Metrics Across SBERT Layers}
\resizebox{\textwidth}{!}{
\begin{tabular}{|c|c|c|c|c|c|c|}
\hline
\textbf{Layer} & \textbf{Mean Sim (Real)} & \textbf{Mean Sim (Permuted)} & \textbf{Norm. Cosine Sim} & \textbf{Hits@5 Acc (Real)} & \textbf{Hits@5 Acc (Permuted)} & \textbf{Norm. Retrieval Acc} \\ \hline
0 & 0.8889 & 0.7808 & 0.4930 & 0.59 & 0.0397 & 0.5731 \\ \hline
1 & 0.9366 & 0.8671 & 0.5228 & 0.57 & 0.0406 & 0.5518 \\ \hline
2 & 0.9397 & 0.8576 & 0.5767 & 0.61 & 0.0390 & 0.5942 \\ \hline
3 & 0.9403 & 0.8330 & 0.6424 & 0.64 & 0.0412 & 0.6245 \\ \hline
4 & 0.9408 & 0.8298 & 0.6523 & 0.66 & 0.0400 & 0.6458 \\ \hline
5 & 0.9409 & 0.8273 & 0.6577 & 0.62 & 0.0417 & 0.6035 \\ \hline
6 & 0.7761 & 0.4865 & 0.5640 & 0.59 & 0.0405 & 0.5727 \\ \hline
\end{tabular}
\label{tab:Compositionality across Layers_appx}
}
\end{table*}

\begin{table*}[t]
\centering
\caption{Additive Compositionality Metrics at Different Training Steps of BERT}
\label{tab:bert_training_stages_appx}
\resizebox{\textwidth}{!}{
\begin{tabular}{|l|c|c|c|c|c|c|c|}
\hline
\textbf{Model} & \textbf{Training Steps} & \textbf{Mean Sim (Real)} & \textbf{Mean Sim (Permuted)} & \textbf{Norm. Cosine Sim} & \textbf{Hits@5 Acc (Real)} & \textbf{Permuted Acc} & \textbf{Norm. Retrieval Acc} \\ \hline
\textbf{cls\_0k}     & 0         & 0.9884 & 0.9882 & 0.0163 & 0.44 & 0.0418 & 0.4156 \\ \hline
\textbf{cls\_20k}    & 20,000    & 0.8787 & 0.7722 & 0.4676 & 0.55 & 0.0407 & 0.5309 \\ \hline
\textbf{cls\_40k}    & 40,000    & 0.8773 & 0.7724 & 0.4607 & 0.48 & 0.0405 & 0.4581 \\ \hline
\textbf{cls\_100k}   & 100,000   & 0.9201 & 0.8323 & 0.5236 & 0.55 & 0.0417 & 0.5304 \\ \hline
\textbf{cls\_1000k}  & 1,000,000 & 0.9545 & 0.9149 & 0.4655 & 0.44 & 0.0408 & 0.4162 \\ \hline
\textbf{cls\_2000k}  & 2,000,000 & 0.9538 & 0.9094 & 0.4896 & 0.48 & 0.0415 & 0.4575 \\ \hline
\end{tabular}
}
\end{table*}




\section{Word Embeddings}
\subsection{WordNet Embedding}
WordNet \citep{miller1995wordnet} is a large lexical database of English that combines dictionary and thesaurus features with a graph structure. It organizes nouns, verbs, adjectives, and adverbs into synsets (sets of cognitive synonyms) interlinked by semantic and lexical relations. We use WN18RR \citep{dettmers2018convolutional}, a subset of WordNet with 40,943 entities and 11 types of relation. For linearity experiments, we select 25,781 words from the intersection of the WN18RR and word2vec (GoogleNews-vectors-negative300) vocabularies. We train embeddings for these words over WordNet on an entity prediction task, which involves predicting the tail entity given a head entity and relation. Training details are provided in Appendix \ref{appendix: WordNet}. We build a continuous embedding matrix $\mathbf{U}_{w2v-wn}\in \mathbb{R}^{25,781\times 300}$ using vectors from GoogleNews-vectors-negative300, and a second continuous embedding matrix $\mathbf{U}_{wn-w2v}\in \mathbb{R}^{25,781\times 20}$.
\label{appendix: WordNet}
We want to ensure our WordNet embedding can contain the semantic relation in it. Therefore, we train the embedding with the task of predicting the tail entity given a head entity and relation. For example, we might want to predict the hypernym of cat:
$$< \textbf{cat}, hypernym, \textbf{?}>$$

\paragraph{Mapping Freebase ID to text}
WordNet is constructed with Freebase ID only, an example triple could be <00260881, hypernym, 00260622>. We follow \citet{villmow2019} to preprocess the data and map each entity with the text with a real meaning. 

The above triple can then be processed with the real semantic meaning: <land reform, hypernym, reform>. The word2vec word embedding is pretrained from a google news corpus. 
We train the WordNet Embedding in the following way:
\begin{enumerate}
    \item We split our dataset to use 90\% for training, 10\% for testing.
    \item Triples of $\left(head, relation, tail\right)$ are encoded as relational triples $\left(h, r, t\right)$.
    \item We randomly initialize embeddings for each $h_i$, $r_j$, $ t_k$, use the scoring function in Equation \ref{eq: transe} and minimize the loss by Margin Loss.
    \item We sampled 20 corrupted entities. Learning rate is set at 0.05 and training epoch at 300. 
\end{enumerate}
Results can be found in the Table \ref{tab:wordnet}, which shows that our WordNet embeddings do contain semantic information.

\begin{table}[h]
\centering
\caption{Link prediction performance for WordNet}
\label{tab:wordnet}
\begin{tabular}{@{}lllcc@{}}
\toprule
 & Hits@1 & Hits@3 & Hits@10 & MRR \\ 
\midrule
WordNet & 0.39 & 0.41 & 0.43 & 0.40 \\  
\bottomrule
\end{tabular}
\end{table}

\section{List of words of experiments: Decomposing word2vec Embedding by Additive Compositionality Detection}
\label{appendix:278words}
allegorically, whimsicality, whimsically, voyeuristically, weightier, weightiest, weightily, weightiness, weightlessly, veritably, visualizations, tyrannically, traitorously, transcendentally, transitionally, tangentially, temperamentally, surgically, structurally, studiously, studiousness, stylistically, spiritualistic, slipperiest, slipperiness, serviceability, serviceably, sectionalism, sentimentalism, sentimentalist, sentimentalized, sentimentalizes, sentimentalizing, sentimentally, serialization, serializations, satanically, reverentially, ritualistically, regularization, quizzically, rapturously, puritanically, probationers, probationer, psychiatrically, preferentially, practicably, practicalities, pleasurably, polarizations, phenomenally, personalizations, pessimistically, pathetically, occupationally, optionally, oratorically, nationalizations, nautically, neutralizer, neutralizers, mysteriously, mysteriousness, narcissistically, moralistically, melodiously, melodiousness, memorializes, memorializing, metrication, materialistically, mechanistically, mechanizations, longitudinally, lexically, liberalization, liquidator, liquidators, journalistic, inferentially, injuriously, hysterically, idealistically, heretically, futuristically, fractionally, fluoridation, fictionalized, fictionalizes, fictionalizing, figuratively, farcically, fatalistic, environmentalists, environmentally, episodically, equitably, emotionlessly, ecclesiastically, editorialized, editorializes, editorializing, editorially, educationalist, educationalists, educationally, egotistical, egotistically, dictatorships, differentiations, derivatively, developmentally, deviationist, deviationists, decoratively, deferentially, definitively, demagogically, demonically, decimalization, cumulatively, conversationalists, conversationally, confidentialities, conspiratorially, collectivization, colonialists, classically, chauvinistically, censoriousness, certifications, capitalizations, breathalyser, breathalysers, brutalization, antagonistically, apocalyptically, weightlessness, westernization, victoriously, visualization, vocalization, urbanization, Unitarian, Unitarians, transcendentalism, transcendentalist, transcendentalists, theatrically, theoretically, technicalities, technicality, technically, speculatively, socialistic, socialization, sophisticate, sophisticated, sophisticates, significantly, sensational, sentimentalists, sentimentality, sentimentalize, scientifically, satirically, rotationally, residentially, relativistic, realistically, prudentially, pressurization, probabilistic, probationary, professionalism, professionally, popularization, potentialities, potentiality, potentially, practicability, practicality, practically, polarization, phosphorescence, phosphorescent, physically, physicalness, periodically, personalization, particularistic, paternalistic, operational, oratorical, organizational, normalization, numerically, negatively, negativism, neutralization, nationalistic, nationalization, naturalistic, naturalization, mechanically, memorialize, memorialized, metrically, maturational, localization, linguistically, liquidation, liquidations, juridical, justifiably, industrialization, imperialistic, incidentally, identically, imaginatively, historically, harmoniously, graphically, generalization, generalizations, fictionalize, existentialism, existentialist, existentialists, evangelicalism, environmentalism, environmentalist, equalization, equalizers, equatorial, electrically, electronically, emotionalism, emotionality, emotionally, economically, editorialist, editorialize, directionality, directionally, dictatorial, dictatorship, differentiation, conversationalist, confidentiality, confidentially, colonialism, colonialist, commercialization, communicational, civilizational, centralization, certification, chemically, capitalistic, capitalization, catastrophically, categorically, behaviorally, authentication, authentications, authenticator, artistically, architecturally, anatomically, Anglicanism, alternatively, altruistically, adventurously, acoustically, activation, additionally

\section{Movie-Lens Training Details}
\label{appendix:movielenstraining}
This experiment was conducted on the MovieLens 1M dataset \citep{harper2015movielens} which consists of a large set of movies and users, and a set of movie ratings for each individual user. It is widely used to create and test recommender systems. Typically, the goal of a recommender system is to predict the rating of an unrated movie for a given user, based on the rest of the data. The dataset contains  6040 users and approximately 3900 movies. Each user-movie rating can take values in 1 to 5.
There are 1 million triples (out of a possible $6040\times3900 = 23.6m$), so that the vast majority of user-movie pairs are not rated. 

Users and movies each have additional attributes attached. For example, users have demographic information such as gender, age, or occupation. Whilst this information is typically used to improve the accuracy of recommendations, we use it to test whether the embedding of a user correlates to private attributes, such as gender or age. We compute our graph embedding based only on ratings, leaving user attributes out. Experiments for training knowledge graph embeddings are implemented with the OpenKE \citep{han-etal-2018-openke} toolkit. We train our model on GeForce GTX TITAN X.

We embed the knowledge graph in the following way:
\begin{enumerate}
    \item We split our dataset to use 90\% for training, 10\% for testing.
    \item Triples of $\left( user, rating, movie\right)$ are encoded as relational triples $\left(h, r, t\right)$.
    \item We randomly initialize embeddings for each $h_i$, $r_j$, $ t_k$ and train embeddings to minimize the loss in \eqref{eq:actual_loss} \footnote{We add a negative sign for the additive scoring function, since we want to maximize the probability of the true triple, which aligns the setting of this loss function}.
    \item We sampled 10 corrupted entities and 4 corrupted relations per true triple. Learning rate is set at 0.01 and training epoch at 300. 
\end{enumerate}
We verify the quality of the embeddings by carrying out a link prediction task on the remaining 10\% test set. We achieved a RMSE score of 0.88, Hits@1 score of 0.46 and Hits@3 as 0.92, MRR as 0.68 and MR as 1.89.

\section{Analysis of the Canonical Component of CCA for Knowledge Graph Embedding}


\begin{figure}[htbp]
    \centering
    \includegraphics[width=\columnwidth]{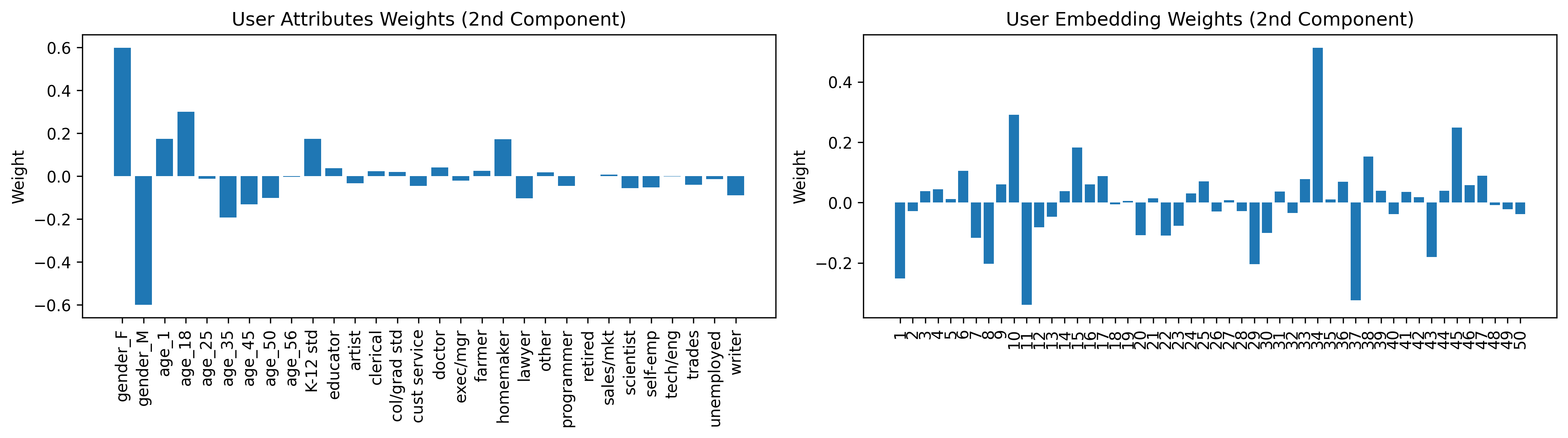}
     \caption{Positive weights in the second canonical component of CCA show the association between user attributes (e.g., \textit{Female} and \textit{Age: 18}) and user embedding dimensions (e.g., \textit{34} and \textit{10}). }
     \label{fig: user-weights}
\end{figure}

\end{document}

%% file: math_commands.tex
\usepackage{amsmath,amsfonts,bm}

\def\eqref#1{equation~\ref{#1}}

\def\1{\bm{1}}

\DeclareMathAlphabet{\mathsfit}{\encodingdefault}{\sfdefault}{m}{sl}
\SetMathAlphabet{\mathsfit}{bold}{\encodingdefault}{\sfdefault}{bx}{n}













%% file: main.bbl
\begin{thebibliography}{61}
\providecommand{\natexlab}[1]{#1}

\bibitem[{Andreas(2019)}]{andreas2019measuring}
Jacob Andreas. 2019.
\newblock Measuring compositionality in representation learning.
\newblock \emph{arXiv preprint arXiv:1902.07181}.

\bibitem[{Berg et~al.(2017)Berg, Kipf, and Welling}]{berg2017graph}
Rianne van~den Berg, Thomas~N Kipf, and Max Welling. 2017.
\newblock Graph convolutional matrix completion.
\newblock \emph{arXiv preprint arXiv:1706.02263}.

\bibitem[{Bordes et~al.(2013)Bordes, Usunier, Garcia-Duran, Weston, and
  Yakhnenko}]{bordes2013translating}
Antoine Bordes, Nicolas Usunier, Alberto Garcia-Duran, Jason Weston, and Oksana
  Yakhnenko. 2013.
\newblock Translating embeddings for modeling multi-relational data.
\newblock \emph{Advances in neural information processing systems}, 26.

\bibitem[{Bose and Hamilton(2019)}]{bose2019compositional}
Avishek Bose and William Hamilton. 2019.
\newblock Compositional fairness constraints for graph embeddings.
\newblock In \emph{International Conference on Machine Learning}, pages
  715--724. PMLR.

\bibitem[{Bowman et~al.(2016)Bowman, Vilnis, Vinyals, Dai, Jozefowicz, and
  Bengio}]{bowman2016generating}
Samuel Bowman, Luke Vilnis, Oriol Vinyals, Andrew Dai, Rafal Jozefowicz, and
  Samy Bengio. 2016.
\newblock Generating sentences from a continuous space.
\newblock In \emph{Proceedings of the 20th SIGNLL Conference on Computational
  Natural Language Learning}, pages 10--21.

\bibitem[{Brown(2020)}]{brown2020language}
Tom~B Brown. 2020.
\newblock Language models are few-shot learners.
\newblock \emph{arXiv preprint arXiv:2005.14165}.

\bibitem[{Chrupa{\l}a and Alishahi(2019)}]{chrupala2019correlating}
Grzegorz Chrupa{\l}a and Afra Alishahi. 2019.
\newblock Correlating neural and symbolic representations of language.
\newblock \emph{arXiv preprint arXiv:1905.06401}.

\bibitem[{Church(2017)}]{church2017word2vec}
Kenneth~Ward Church. 2017.
\newblock Word2vec.
\newblock \emph{Natural Language Engineering}, 23(1):155--162.

\bibitem[{Cook et~al.(2007)Cook, Fazly, and Stevenson}]{cook2007pulling}
Paul Cook, Afsaneh Fazly, and Suzanne Stevenson. 2007.
\newblock Pulling their weight: Exploiting syntactic forms for the automatic
  identification of idiomatic expressions in context.
\newblock In \emph{Proceedings of the workshop on a broader perspective on
  multiword expressions}, pages 41--48.

\bibitem[{Dalpiaz et~al.(2018)Dalpiaz, Van~der Schalk, and
  Lucassen}]{dalpiaz2018pinpointing}
Fabiano Dalpiaz, Ivor Van~der Schalk, and Garm Lucassen. 2018.
\newblock Pinpointing ambiguity and incompleteness in requirements engineering
  via information visualization and nlp.
\newblock In \emph{Requirements Engineering: Foundation for Software Quality:
  24th International Working Conference, REFSQ 2018, Utrecht, The Netherlands,
  March 19-22, 2018, Proceedings 24}, pages 119--135. Springer.

\bibitem[{Dankers and Lucas(2023)}]{dankers2023non}
Verna Dankers and Christopher Lucas. 2023.
\newblock Non-compositionality in sentiment: New data and analyses.
\newblock In \emph{Findings of the Association for Computational Linguistics:
  EMNLP 2023}, pages 5150--5162.

\bibitem[{Dettmers et~al.(2018)Dettmers, Minervini, Stenetorp, and
  Riedel}]{dettmers2018convolutional}
Tim Dettmers, Pasquale Minervini, Pontus Stenetorp, and Sebastian Riedel. 2018.
\newblock Convolutional 2d knowledge graph embeddings.
\newblock In \emph{Proceedings of the AAAI conference on artificial
  intelligence}, volume~32.

\bibitem[{Devlin et~al.(2018)Devlin, Chang, Lee, and
  Toutanova}]{devlin2018bert}
Jacob Devlin, Ming-Wei Chang, Kenton Lee, and Kristina Toutanova. 2018.
\newblock Bert: Pre-training of deep bidirectional transformers for language
  understanding.
\newblock ArXiv preprint arXiv:1810.04805.

\bibitem[{Elmoznino et~al.(2024)Elmoznino, Jiralerspong, Bengio, and
  Lajoie}]{elmoznino2024complexity}
Eric Elmoznino, Thomas Jiralerspong, Yoshua Bengio, and Guillaume Lajoie. 2024.
\newblock A complexity-based theory of compositionality.
\newblock \emph{arXiv preprint arXiv:2410.14817}.

\bibitem[{Ettinger et~al.(2016)Ettinger, Elgohary, and
  Resnik}]{ettinger2016probing}
Allyson Ettinger, Ahmed Elgohary, and Philip Resnik. 2016.
\newblock Probing for semantic evidence of composition by means of simple
  classification tasks.
\newblock In \emph{Proceedings of the 1st workshop on evaluating vector-space
  representations for nlp}, pages 134--139.

\bibitem[{Fournier et~al.(2020)Fournier, Dupoux, and
  Dunbar}]{fournier2020analogies}
Louis Fournier, Emmanuel Dupoux, and Ewan Dunbar. 2020.
\newblock Analogies minus analogy test: measuring regularities in word
  embeddings.
\newblock In \emph{CoNLL 2020-24th Conference on Computational Natural Language
  Learning}.

\bibitem[{Gittens et~al.(2017)Gittens, Achlioptas, and
  Mahoney}]{gittens2017skip}
Alex Gittens, Dimitris Achlioptas, and Michael~W Mahoney. 2017.
\newblock Skip-gram- zipf+ uniform= vector additivity.
\newblock In \emph{Proceedings of the 55th Annual Meeting of the Association
  for Computational Linguistics (Volume 1: Long Papers)}, pages 69--76.

\bibitem[{Goldberg and Levy(2014)}]{goldberg2014word2vec}
Yoav Goldberg and Omer Levy. 2014.
\newblock word2vec explained: deriving mikolov et al.'s negative-sampling
  word-embedding method.
\newblock \emph{arXiv preprint arXiv:1402.3722}.

\bibitem[{Guo and Wu(2019)}]{guo2019canonical}
Chenfeng Guo and Dongrui Wu. 2019.
\newblock Canonical correlation analysis (cca) based multi-view learning: An
  overview.
\newblock \emph{arXiv preprint arXiv:1907.01693}.

\bibitem[{Guo et~al.(2023)Guo, Xu, Lewis, and Cristianini}]{Guo2023}
Zhijin Guo, Zhaozhen Xu, Martha Lewis, and Nello Cristianini. 2023.
\newblock \href {https://aequitas-aod.github.io/aequitas-ecai23.github.io/}
  {Extract: Explainable transparent control of bias in embeddings}.
\newblock In \emph{Proceedings of the 1st Workshop on Fairness and Bias in AI
  co-located with 26th European Conference on Artificial Intelligence (ECAI
  2023)}.

\bibitem[{Han et~al.(2018)Han, Cao, Lv, Lin, Liu, Sun, and
  Li}]{han-etal-2018-openke}
Xu~Han, Shulin Cao, Xin Lv, Yankai Lin, Zhiyuan Liu, Maosong Sun, and Juanzi
  Li. 2018.
\newblock \href {https://doi.org/10.18653/v1/D18-2024} {{O}pen{KE}: An open
  toolkit for knowledge embedding}.
\newblock In \emph{Proceedings of the 2018 Conference on Empirical Methods in
  Natural Language Processing: System Demonstrations}, pages 139--144,
  Brussels, Belgium. Association for Computational Linguistics.

\bibitem[{Harper and Konstan(2015)}]{harper2015movielens}
F~Maxwell Harper and Joseph~A Konstan. 2015.
\newblock The movielens datasets: History and context.
\newblock \emph{Acm transactions on interactive intelligent systems (tiis)},
  5(4):1--19.

\bibitem[{Hernandez et~al.(2024)Hernandez, Sharma, Haklay, Meng, Wattenberg,
  Andreas, Belinkov, and Bau}]{hernandezlinearity}
Evan Hernandez, Arnab~Sen Sharma, Tal Haklay, Kevin Meng, Martin Wattenberg,
  Jacob Andreas, Yonatan Belinkov, and David Bau. 2024.
\newblock Linearity of relation decoding in transformer language models.
\newblock In \emph{The Twelfth International Conference on Learning
  Representations}.

\bibitem[{Hewitt and Manning(2019)}]{hewitt2019structural}
John Hewitt and Christopher~D Manning. 2019.
\newblock A structural probe for finding syntax in word representations.
\newblock In \emph{Proceedings of the 2019 Conference of the North American
  Chapter of the Association for Computational Linguistics: Human Language
  Technologies, Volume 1 (Long and Short Papers)}, pages 4129--4138.

\bibitem[{Hofmann et~al.(2025)Hofmann, Weissweiler, Mortensen, Sch{\"u}tze, and
  Pierrehumbert}]{hofmann2025derivational}
Valentin Hofmann, Leonie Weissweiler, David~R Mortensen, Hinrich Sch{\"u}tze,
  and Janet~B Pierrehumbert. 2025.
\newblock Derivational morphology reveals analogical generalization in large
  language models.
\newblock \emph{Proceedings of the National Academy of Sciences},
  122(19):e2423232122.

\bibitem[{Jackendoff et~al.(2011)Jackendoff, Lieber, and
  Štekauer}]{JackendoffRay2011CitP}
Ray Jackendoff, Rochelle Lieber, and Pavol Štekauer. 2011.
\newblock Compounding in the parallel architecture and conceptual semantics.
\newblock In \emph{The Oxford Handbook of Compounding}, Oxford Handbooks in
  Linguistics. Oxford University Press.

\bibitem[{Kumar et~al.(2019)Kumar, Pujari, Padmanabhan, and
  Kagita}]{kumar2019group}
Vikas Kumar, Arun~K Pujari, Vineet Padmanabhan, and Venkateswara~Rao Kagita.
  2019.
\newblock Group preserving label embedding for multi-label classification.
\newblock \emph{Pattern Recognition}, 90:23--34.

\bibitem[{Lepori and McCoy(2020)}]{lepori-mccoy-2020-picking}
Michael Lepori and R.~Thomas McCoy. 2020.
\newblock \href {https://doi.org/10.18653/v1/2020.coling-main.325} {Picking
  {BERT}{'}s brain: Probing for linguistic dependencies in contextualized
  embeddings using representational similarity analysis}.
\newblock In \emph{Proceedings of the 28th International Conference on
  Computational Linguistics}, pages 3637--3651, Barcelona, Spain (Online).
  International Committee on Computational Linguistics.

\bibitem[{Lepori et~al.(2023)Lepori, Serre, and Pavlick}]{lepori2023break}
Michael Lepori, Thomas Serre, and Ellie Pavlick. 2023.
\newblock Break it down: Evidence for structural compositionality in neural
  networks.
\newblock \emph{Advances in Neural Information Processing Systems},
  36:42623--42660.

\bibitem[{Levy and Goldberg(2014)}]{levy2014neural}
Omer Levy and Yoav Goldberg. 2014.
\newblock Neural word embedding as implicit matrix factorization.
\newblock \emph{Advances in neural information processing systems}, 27.

\bibitem[{Li et~al.(2020)Li, Zhou, He, Wang, Yang, and Li}]{li2020sentence}
Bohan Li, Hao Zhou, Junxian He, Mingxuan Wang, Yiming Yang, and Lei Li. 2020.
\newblock On the sentence embeddings from pre-trained language models.
\newblock In \emph{Proceedings of the 2020 Conference on Empirical Methods in
  Natural Language Processing (EMNLP)}, pages 9119--9130.

\bibitem[{Lin(1999)}]{lin1999automatic}
Dekang Lin. 1999.
\newblock Automatic identification of non-compositional phrases.
\newblock In \emph{Proceedings of the 37th annual meeting of the Association
  for Computational Linguistics}, pages 317--324.

\bibitem[{McCarthy et~al.(2007)McCarthy, Venkatapathy, and
  Joshi}]{mccarthy2007detecting}
Diana McCarthy, Sriram Venkatapathy, and Aravind Joshi. 2007.
\newblock Detecting compositionality of verb-object combinations using
  selectional preferences.
\newblock In \emph{Proceedings of the 2007 Joint Conference on Empirical
  Methods in Natural Language Processing and Computational Natural Language
  Learning (EMNLP-CoNLL)}, pages 369--379.

\bibitem[{Meng et~al.(2022)Meng, Bau, Andonian, and
  Belinkov}]{meng2022locating}
Kevin Meng, David Bau, Alex Andonian, and Yonatan Belinkov. 2022.
\newblock Locating and editing factual associations in gpt.
\newblock \emph{Advances in neural information processing systems},
  35:17359--17372.

\bibitem[{Mikolov et~al.(2013{\natexlab{a}})Mikolov, Chen, Corrado, and
  Dean}]{mikolov2013efficient}
Tomas Mikolov, Kai Chen, Greg Corrado, and Jeffrey Dean. 2013{\natexlab{a}}.
\newblock Efficient estimation of word representations in vector space.
\newblock \emph{arXiv preprint arXiv:1301.3781}.

\bibitem[{Mikolov et~al.(2013{\natexlab{b}})Mikolov, Sutskever, Chen, Corrado,
  and Dean}]{mikolov2013distributed}
Tomas Mikolov, Ilya Sutskever, Kai Chen, Greg~S Corrado, and Jeff Dean.
  2013{\natexlab{b}}.
\newblock Distributed representations of words and phrases and their
  compositionality.
\newblock \emph{Advances in neural information processing systems}, 26.

\bibitem[{Mikolov et~al.(2013{\natexlab{c}})Mikolov, Yih, and
  Zweig}]{mikolov2013linguistic}
Tom{\'a}{\v{s}} Mikolov, Wen-tau Yih, and Geoffrey Zweig. 2013{\natexlab{c}}.
\newblock Linguistic regularities in continuous space word representations.
\newblock In \emph{Proceedings of the 2013 conference of the north american
  chapter of the association for computational linguistics: Human language
  technologies}, pages 746--751.

\bibitem[{Miller(1995)}]{miller1995wordnet}
George~A Miller. 1995.
\newblock Wordnet: a lexical database for english.
\newblock \emph{Communications of the ACM}, 38(11):39--41.

\bibitem[{Nickel et~al.(2011)Nickel, Tresp, and Kriegel}]{nickel2011three}
Maximilian Nickel, Volker Tresp, and Hans-Peter Kriegel. 2011.
\newblock A three-way model for collective learning on multi-relational data.
\newblock In \emph{Icml}.

\bibitem[{Oyama et~al.(2024)Oyama, Yamagiwa, and
  Shimodaira}]{oyama2024understanding}
Momose Oyama, Hiroaki Yamagiwa, and Hidetoshi Shimodaira. 2024.
\newblock Understanding higher-order correlations among semantic components in
  embeddings.
\newblock In \emph{Proceedings of the 2024 Conference on Empirical Methods in
  Natural Language Processing}, pages 2883--2899.

\bibitem[{Park et~al.(2024)Park, Choe, and Veitch}]{park2024linear}
Kiho Park, Yo~Joong Choe, and Victor Veitch. 2024.
\newblock The linear representation hypothesis and the geometry of large
  language models.
\newblock In \emph{International Conference on Machine Learning}, pages
  39643--39666. PMLR.

\bibitem[{Rastogi et~al.(2020)Rastogi, Zang, Sunkara, Gupta, and
  Khaitan}]{rastogi2020towards}
Abhinav Rastogi, Xiaoxue Zang, Srinivas Sunkara, Raghav Gupta, and Pranav
  Khaitan. 2020.
\newblock Towards scalable multi-domain conversational agents: The
  schema-guided dialogue dataset.
\newblock In \emph{Proceedings of the AAAI Conference on Artificial
  Intelligence}, volume~34, pages 8689--8696.

\bibitem[{S{\'a}nchez-Guti{\'e}rrez et~al.(2018)S{\'a}nchez-Guti{\'e}rrez,
  Mailhot, Deacon, and Wilson}]{sanchez2018morpholex}
Claudia~H S{\'a}nchez-Guti{\'e}rrez, Hugo Mailhot, S~H{\'e}l{\`e}ne Deacon, and
  Maximiliano~A Wilson. 2018.
\newblock Morpholex: A derivational morphological database for 70,000 english
  words.
\newblock \emph{Behavior research methods}, 50:1568--1580.

\bibitem[{Sellam et~al.(2022)Sellam, Yadlowsky, Tenney, Wei, Saphra, D'Amour,
  Linzen, Bastings, Turc, Eisenstein et~al.}]{sellam2022multiberts}
Thibault Sellam, Steve Yadlowsky, Ian Tenney, Jason Wei, Naomi Saphra,
  Alexander D'Amour, Tal Linzen, Jasmijn Bastings, Iulia Turc, Jacob
  Eisenstein, and 1 others. 2022.
\newblock The multiberts: Bert reproductions for robustness analysis.
\newblock In \emph{10th International Conference on Learning Representations,
  ICLR 2022}.

\bibitem[{Seoane et~al.(2014)Seoane, Campbell, Day, Casas, and
  Gaunt}]{seoane2014canonical}
Jose~A Seoane, Colin Campbell, Ian~NM Day, Juan~P Casas, and Tom~R Gaunt. 2014.
\newblock Canonical correlation analysis for gene-based pleiotropy discovery.
\newblock \emph{PLoS computational biology}, 10(10):e1003876.

\bibitem[{Seonwoo et~al.(2019)Seonwoo, Park, Kim, and Oh}]{seonwoo2019additive}
Yeon Seonwoo, Sungjoon Park, Dongkwan Kim, and Alice Oh. 2019.
\newblock Additive compositionality of word vectors.
\newblock In \emph{Proceedings of the 5th Workshop on Noisy User-generated Text
  (W-NUT 2019)}, pages 387--396.

\bibitem[{Shawe-Taylor et~al.(2004)Shawe-Taylor, Cristianini
  et~al.}]{shawe2004kernel}
John Shawe-Taylor, Nello Cristianini, and 1 others. 2004.
\newblock \emph{Kernel methods for pattern analysis}.
\newblock Cambridge university press.

\bibitem[{Tapanainen et~al.(1998)Tapanainen, Piitulainen, and
  Jarvinen}]{tapanainen1998idiomatic}
Pasi Tapanainen, Jussi Piitulainen, and Timo Jarvinen. 1998.
\newblock Idiomatic object usage and support verbs.
\newblock In \emph{COLING 1998 Volume 2: The 17th International Conference on
  Computational Linguistics}.

\bibitem[{Tenney et~al.(2019{\natexlab{a}})Tenney, Xia, Chen, Wang, Poliak,
  McCoy, Kim, Van~Durme, Bowman, Das et~al.}]{tenney2019you}
Ian Tenney, Patrick Xia, Berlin Chen, Alex Wang, Adam Poliak, R~Thomas McCoy,
  Najoung Kim, Benjamin Van~Durme, Samuel~R Bowman, Dipanjan Das, and 1 others.
  2019{\natexlab{a}}.
\newblock What do you learn from context? probing for sentence structure in
  contextualized word representations.
\newblock \emph{arXiv preprint arXiv:1905.06316}.

\bibitem[{Tenney et~al.(2019{\natexlab{b}})Tenney, Xia, Chen, Wang, Poliak,
  McCoy, Kim, Van~Durme, Bowman, Das et~al.}]{tenney2019}
Ian Tenney, Patrick Xia, Berlin Chen, Alex Wang, Adam Poliak, R~Thomas McCoy,
  Najoung Kim, Benjamin Van~Durme, Samuel~R Bowman, Dipanjan Das, and 1 others.
  2019{\natexlab{b}}.
\newblock What do you learn from context? probing for sentence structure in
  contextualized word representations.
\newblock In \emph{7th International Conference on Learning Representations,
  ICLR 2019}.

\bibitem[{Thilak et~al.(2024)Thilak, Huang, Saremi, Dinh, Goh, Nakkiran,
  Susskind, and Littwin}]{thilaklidar}
Vimal Thilak, Chen Huang, Omid Saremi, Laurent Dinh, Hanlin Goh, Preetum
  Nakkiran, Joshua~M Susskind, and Etai Littwin. 2024.
\newblock Lidar: Sensing linear probing performance in joint embedding ssl
  architectures.
\newblock In \emph{The Twelfth International Conference on Learning
  Representations}.

\bibitem[{Touvron et~al.(2023)Touvron, Lavril, Izacard, Martinet, Lachaux,
  Lacroix, Rozi{\`e}re, Goyal, Hambro, Azhar et~al.}]{touvron2023llama}
Hugo Touvron, Thibaut Lavril, Gautier Izacard, Xavier Martinet, Marie-Anne
  Lachaux, Timoth{\'e}e Lacroix, Baptiste Rozi{\`e}re, Naman Goyal, Eric
  Hambro, Faisal Azhar, and 1 others. 2023.
\newblock Llama: Open and efficient foundation language models.
\newblock \emph{arXiv preprint arXiv:2302.13971}.

\bibitem[{Trager et~al.(2023)Trager, Perera, Zancato, Achille, Bhatia, and
  Soatto}]{trager2023linear}
Matthew Trager, Pramuditha Perera, Luca Zancato, Alessandro Achille, Parminder
  Bhatia, and Stefano Soatto. 2023.
\newblock Linear spaces of meanings: compositional structures in
  vision-language models.
\newblock In \emph{Proceedings of the IEEE/CVF International Conference on
  Computer Vision}, pages 15395--15404.

\bibitem[{Trouillon et~al.(2016)Trouillon, Welbl, Riedel, Gaussier, and
  Bouchard}]{trouillon2016complex}
Th{\'e}o Trouillon, Johannes Welbl, Sebastian Riedel, {\'E}ric Gaussier, and
  Guillaume Bouchard. 2016.
\newblock Complex embeddings for simple link prediction.
\newblock In \emph{International conference on machine learning}, pages
  2071--2080. PMLR.

\bibitem[{villmow(2019)}]{villmow2019}
villmow. 2019.
\newblock \href {https://github.com/villmow/datasets_knowledge_embedding}
  {Datasets for knowledge graph completion with textual information about
  entities}.
\newblock Github.

\bibitem[{Vinokourov et~al.(2002)Vinokourov, Cristianini, and
  Shawe-Taylor}]{NIPS2002_d5e2fbef}
Alexei Vinokourov, Nello Cristianini, and John Shawe-Taylor. 2002.
\newblock \href
  {https://proceedings.neurips.cc/paper/2002/file/d5e2fbef30a4eb668a203060ec8e5eef-Paper.pdf}
  {Inferring a semantic representation of text via cross-language correlation
  analysis}.
\newblock In \emph{Advances in Neural Information Processing Systems},
  volume~15. MIT Press.

\bibitem[{Wang and Xu(2024)}]{wang2024composition}
Tianqi Wang and Xu~Xu. 2024.
\newblock Composition as nonlinear combination in semantic space: Exploring the
  effect of compositionality on chinese compound recognition.
\newblock In \emph{Proceedings of the Annual Meeting of the Cognitive Science
  Society}, volume~46.

\bibitem[{Wei et~al.(2008)Wei, Zhao, and Zhu}]{wei2008ontology}
Shikui Wei, Yao Zhao, and Zhenfeng Zhu. 2008.
\newblock Ontology-based inter-concept relation fusion for concept detection.
\newblock In \emph{Advances in Multimedia Information Processing-PCM 2008: 9th
  Pacific Rim Conference on Multimedia, Tainan, Taiwan, December 9-13, 2008.
  Proceedings 9}, pages 721--730. Springer.

\bibitem[{Xu et~al.(2023)Xu, Guo, and Cristianini}]{xu2023on}
Zhaozhen Xu, Zhijin Guo, and Nello Cristianini. 2023.
\newblock On compositionality in data embedding.
\newblock In \emph{Advances in Intelligent Data Analysis XXI: 21st
  International Symposium, IDA 2023}. Springer.

\bibitem[{Yang et~al.(2014)Yang, Yih, He, Gao, and Deng}]{yang2014embedding}
Bishan Yang, Wen-tau Yih, Xiaodong He, Jianfeng Gao, and Li~Deng. 2014.
\newblock Embedding entities and relations for learning and inference in
  knowledge bases.
\newblock \emph{arXiv preprint arXiv:1412.6575}.

\bibitem[{Yazdani et~al.(2015)Yazdani, Farahmand, and
  Henderson}]{yazdani2015learning}
Majid Yazdani, Meghdad Farahmand, and James Henderson. 2015.
\newblock Learning semantic composition to detect non-compositionality of
  multiword expressions.
\newblock In \emph{Proceedings of the 2015 conference on empirical methods in
  natural language processing}, pages 1733--1742.

\end{thebibliography}
